\documentclass[runningheads]{llncs}
\usepackage{graphicx}
\usepackage{amsmath,amssymb} 
\usepackage{color}
\usepackage{wrapfig}
\usepackage[export]{adjustbox}
\usepackage{caption}
\usepackage{booktabs}
\usepackage[T1]{fontenc}
\usepackage[utf8]{inputenc}
\usepackage{authblk}
\usepackage{hyperref}
\usepackage{lipsum}
\hypersetup{
    colorlinks=true,
    linkcolor=blue,
    filecolor=cyan,      
    urlcolor=magenta,
}
\begin{document}
\pagestyle{headings}
\mainmatter

\def\ACCV20SubNumber{0307}  

\title{TTPLA: An Aerial-Image Dataset for Detection and Segmentation of Transmission Towers and Power Lines} 
\titlerunning{TTPLA Dataset}
%
\author{Rabab Abdelfattah\inst{1}\and
Xiaofeng Wang\inst{1} \and
Song Wang\inst{2}}
\authorrunning{R. Abdelfattah et al.}
%
\institute{Department of Electrical Engineering, University of South Carolina, USA \\
 \and
Department of Computer Science and Engineering, University of South Carolina, USA\\
\email{rabab@email.sc.edu,wangxi@cec.sc.edu,songwang@cec.sc.edu}}

\maketitle


\begin{abstract}
Accurate detection and segmentation of transmission towers~(TTs) and power lines~(PLs) from aerial images plays a key role in protecting power-grid security and low-altitude UAV safety. Meanwhile, aerial images of TTs and PLs pose a number of new challenges to the computer vision researchers who work on object detection and segmentation -- PLs are long and thin, and may show similar color as the background; TTs can be of various shapes and most likely made up of line structures of various sparsity; The background scene, lighting, and object sizes can vary significantly from one image to another. In this paper we collect and release a new TT/PL Aerial-image (TTPLA) dataset, consisting of 1,100 images with the resolution of 3,840$\times$2,160 pixels, as well as manually labeled 8,987 instances of TTs and PLs. We develop novel policies for collecting, annotating, and labeling the images in TTPLA. Different from other relevant datasets, TTPLA supports evaluation of instance segmentation, besides detection and semantic segmentation. To build a baseline for detection and segmentation tasks on TTPLA, we report the performance of several state-of-the-art deep learning models on our dataset. TTPLA dataset is publicly available at \url{https://github.com/r3ab/ttpla_dataset}
\end{abstract}


\section{Introduction}

Power grid monitoring and inspection is extremely important to prevent power failures and potential blackouts.  Traditional methods to inspect transmission towers~(TTs) and power lines~(PLs) include visual surveys by human inspectors, helicopter-assisted inspection~\cite{luque2014power}, and crawling robots~\cite{luque2014power}, to name a few. However, these methods always suffer from their high costs in time, labor, and finance, as well as inspection accuracy. 
%
As an alternative, inspection based on small-scale unmanned aerial vehicles~(UAVs) becomes popular and gradually plays an essential role, thanks to its low costs, high mobility and flexibility, and the potential to obtain high-quality images.

Autonomous UAV-based power grid inspection requires precise scene understanding in real-time to enable UAV localization, scene recognition, tracking, aerial monitoring, inspection, and flight safety. The main challenge in fulfilling this requirement, however, points to \textit{background complexity} and \textit{object complexity}. Background complexity mainly comes from the similarity between the color of the PLs and their backgrounds.  Object complexity can be interpreted from four aspects: ($i$) Scale imbalance -- As discussed in~\cite{hu2018relation}, combining strongly correlated objects, such as TTs and PLs, together can potentially enhance recognition accuracy, compared with recognizing them separately.  However, the scales of TTs and PLs are obviously imbalanced in an image; ($ii$) Class imbalance -- In most cases, each TT is linked to least between 3-4 and up to 10 PLs, which will result in significant imbalance among the number of TTs and PLs~\cite{ouyang2016factors,pang2019libra}; ($iii$) Crowded objects -- PLs are very close to each other and sometimes even overlapped in images~\cite{liu2019adaptive}; and ($iv$) Complicated structures and/or shapes.  

\begin{wrapfigure}{r}{0.45\textwidth}
\vskip-9mm
	\includegraphics[scale=0.04]{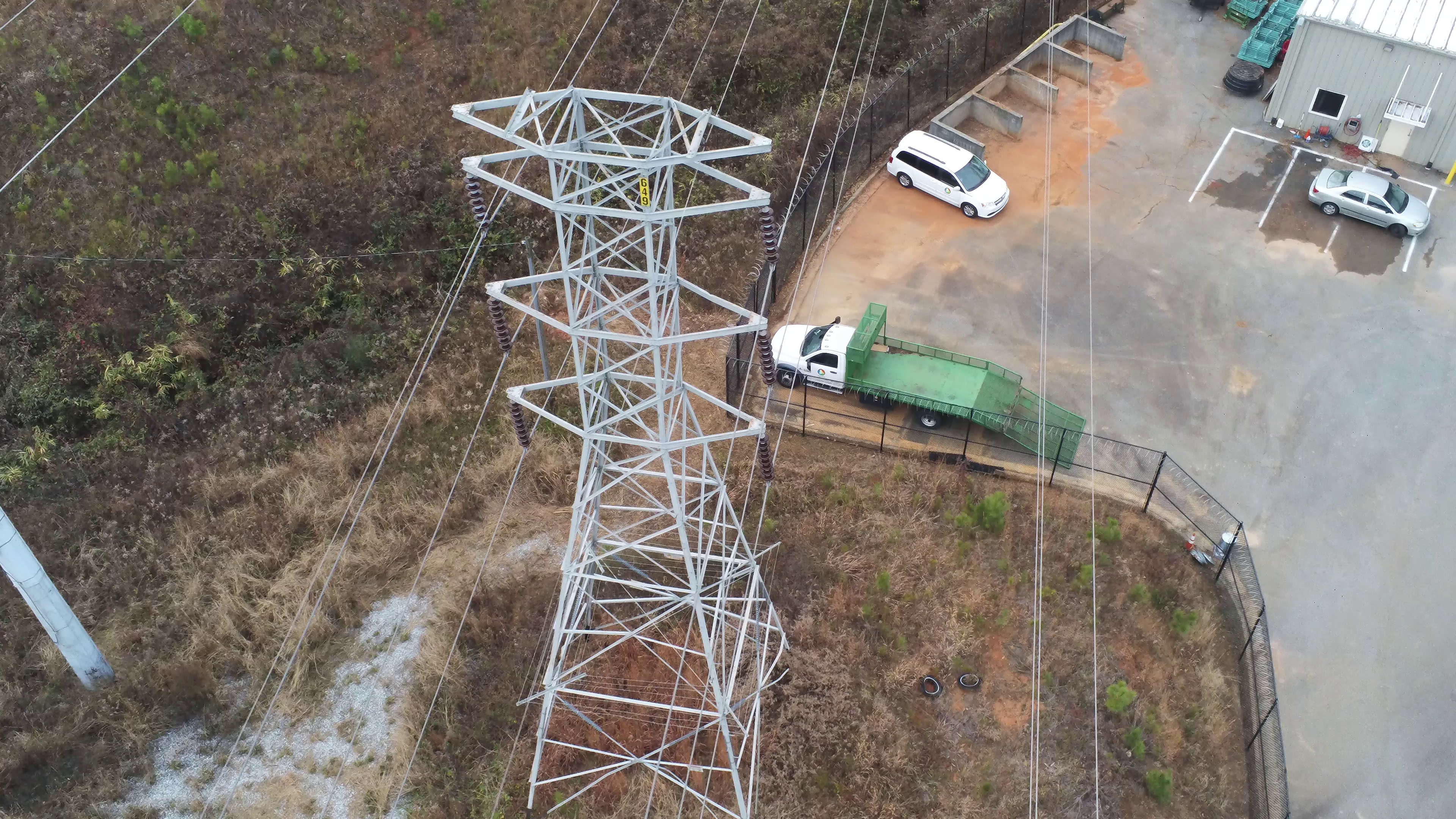} \\
	\vskip-3.5mm
	\includegraphics[scale=0.04]{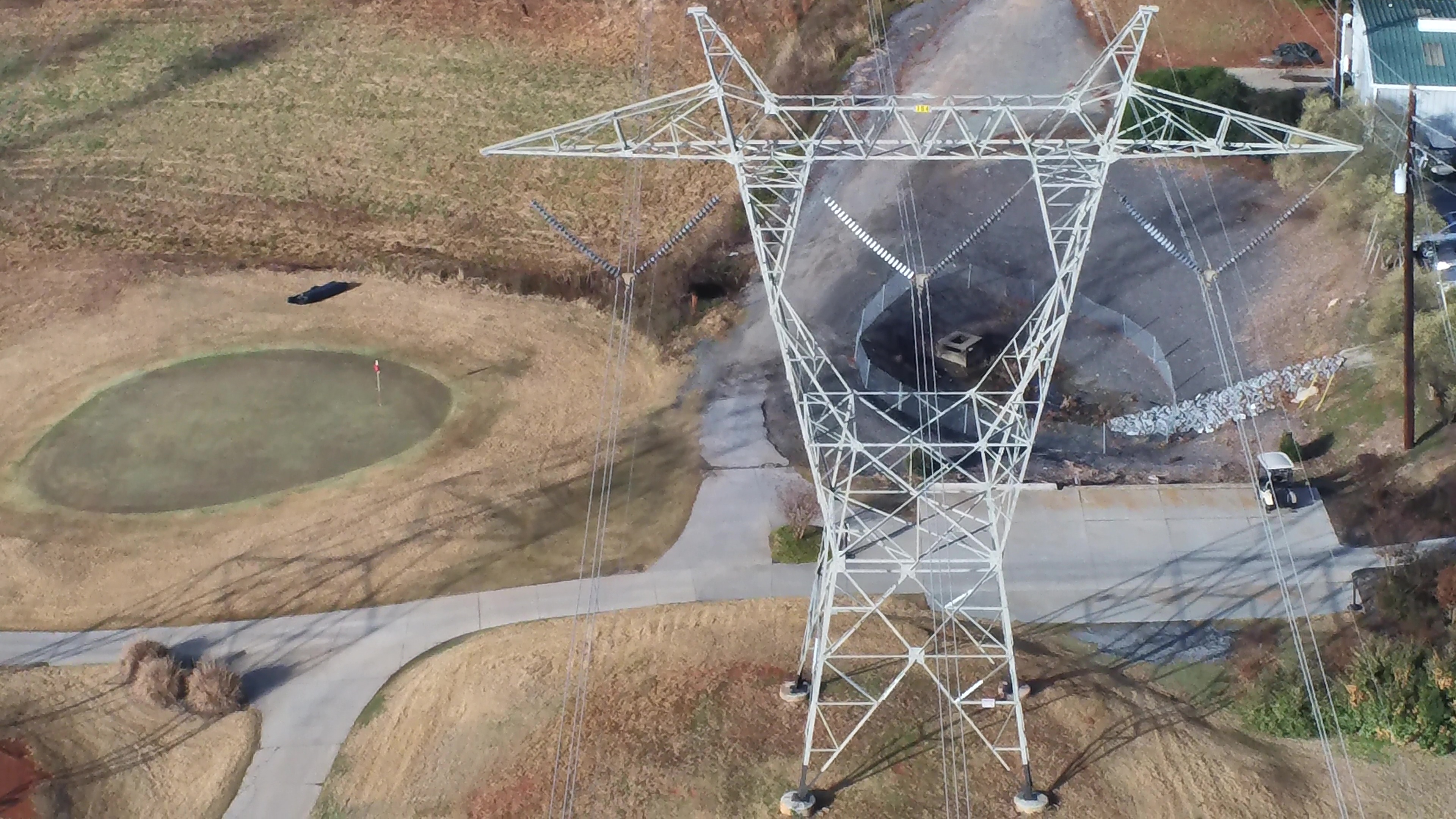}
	\vskip-1mm
	\caption{Different TTs in TTPLA.}
	\label{fig:towershape}
\vskip-6mm
\end{wrapfigure}

PLs are long and thin which makes the distribution of the related pixels in an image completely different from regular objects in some well-known datasets~\cite{lin2014microsoft,cordts2016cityscapes,everingham2010pascal}. Meanwhile, TTs may be of various shapes and most likely made up of line structures of various sparsity, as shown in Fig.~\ref{fig:towershape}.


Existing datasets on TTs and PLs only support two types of annotation: image-level and semantic segmentation~\cite{lin2016efficient,li2017not,papandreou2015weakly,luo2017deep}.  As a result, most related computer vision research only classifies and localizes objects in an image without distinguishing different objects of the same class~\cite{ren2017end} (e.g., it can recognize and/or localize PLs as one class, but cannot distinguish different PLs in the image).  To overcome such a limitation, this paper presents a unique dataset on TTs and PLs, TTPLA~(TT/PL Aerial-image), focusing on a different type of annotation, instance segmentation, which is a combination of object detection and mask segmentation.  
Instance segmentation can distinguish instances (or ``objects'') that belong to the same class and provide a solid understanding for each individual instance.  This is especially useful in power grid inspection, where PLs can be close to each other, occluded, and overlapped.

TTPLA dataset places unique challenges to computer vision research on instance segmentation~\cite{he2017mask,xie2020polarmask,huang2019mask,bolya2019yolact}.  To evaluate TTPLA dataset, we build a baseline on our dataset using Yolact as the instance segmentation model.  The results are collected based on different backbones (Resnet-50 and Resnet-101) and different image resolutions $640\times360$ (preserve aspect ratio), $550\times550$ and $700\times700$. The best average scores for bounding box and mask are $22.96\%$ and $15.72\%$, respectively (more detailed can be found in Table~\ref{results}), which are low in general.  Another observation is that all of the previously mentioned models use the NMS method to sharply filter the large number of false positives near the ground truth~\cite{bodla2017soft,hosang2017learning}, while single NMS may not be practical on our dataset since TTPLA considers a crowded scenario. Therefore, using lower NMS threshold leads to missing highly overlapped objects while using higher NMS threshold leads to increased false positives ~\cite{liu2019adaptive}. Overall, the state-of-the-art approaches may not perform well on TTPLA, which actually motivates the development of novel instance segmentation models.


The main contributions of this paper are described as follows.
\begin{itemize}

\item We present a public dataset, TTPLA, which is a collection of aerial images on TTs and PLs.  The images are taken from different view angles and collected at different time, locations, and backgrounds with different tower structures. 
\item Novel policies are introduced for collecting and labeling images. 
\item Pixel-wise annotation level is chosen to label instances in TTPLA.  This annotations are provided in the COCO format~\cite{lin2014microsoft} which can be easily integrated to other datasets to enrich future research in scene understanding field. To the best of our knowledge, TTPLA is the first public image dataset on TTs and PLs, focusing on instance segmentation, while all the related datasets focus on semantic segmentation.
\item We provide a baseline on TTPLA by evaluating it using the state-of-the-art deep learning models. 
\end{itemize}

    \begin{figure}[t]
	\centering
	\begin{tabular}{c c c c} 
	\includegraphics[width=2.5cm,height=2.8cm]{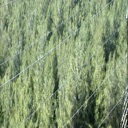}&
	\includegraphics[width=2.5cm,height=2.8cm]{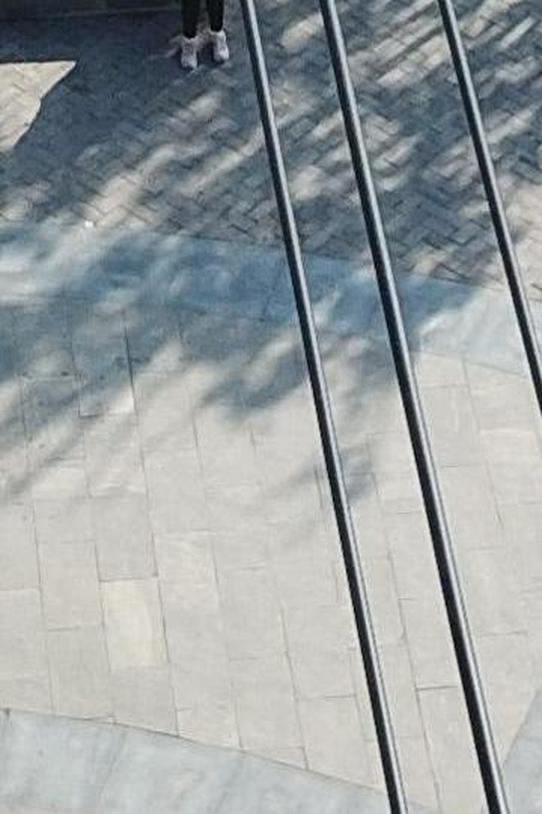}&
	\includegraphics[width=2.5cm,height=2.8cm]{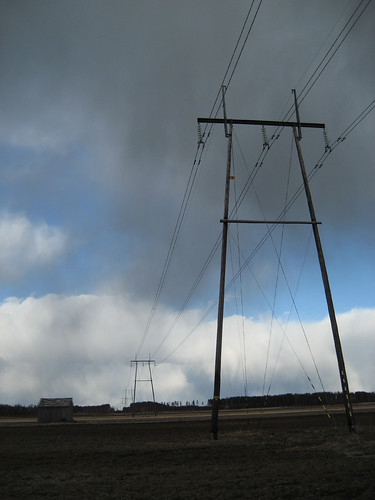}&
	\includegraphics[width=2.5cm,height=2.8cm]{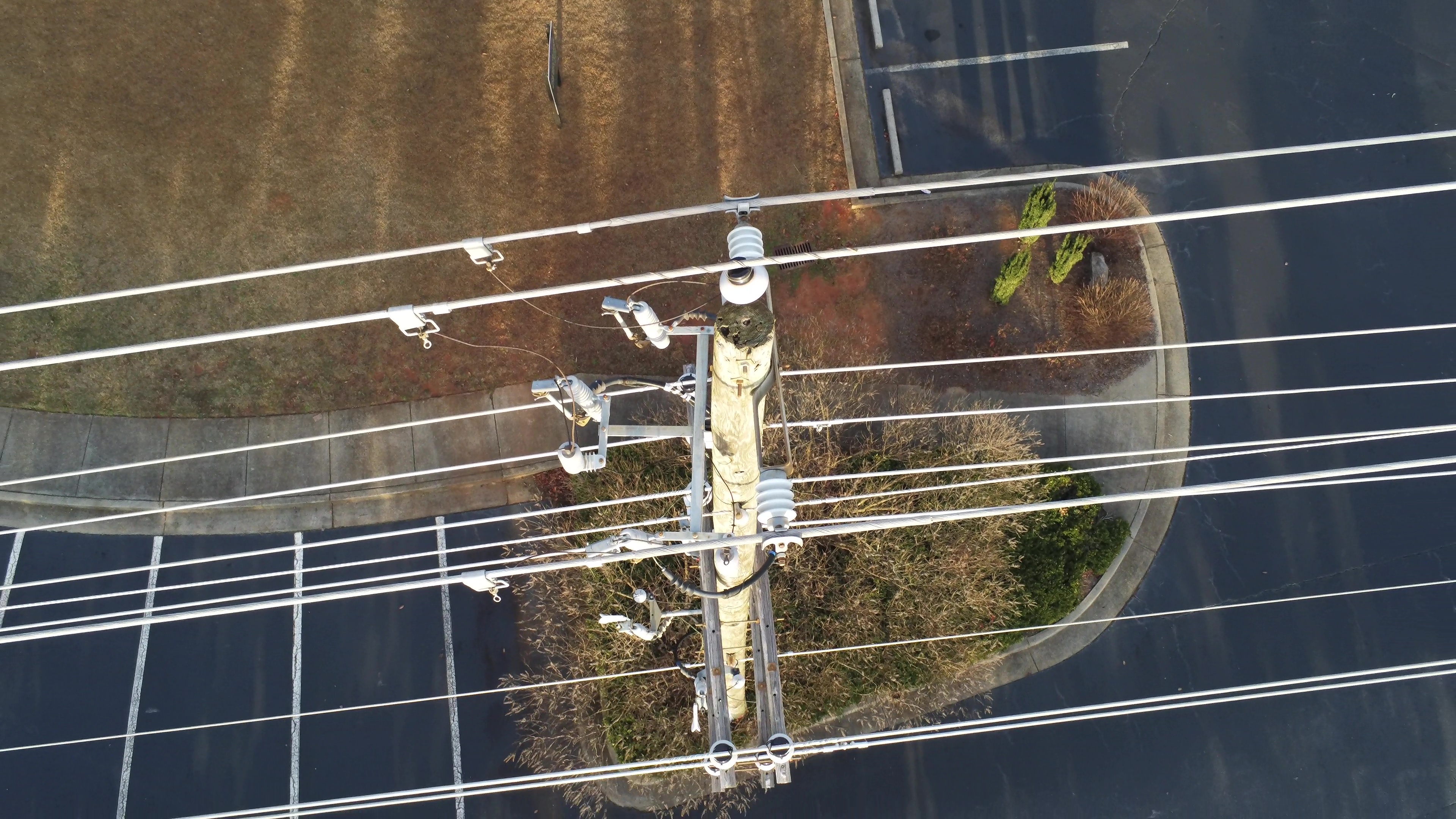}\\
	\footnotesize (a) &
	\footnotesize (b) &
	\footnotesize (c) &
	\footnotesize (d) \\
	\end{tabular}
		\vskip-3.5mm
	\caption{Sample images from public datasets as compared to our dataset TTPLA where  (a)  dataset \cite{emre2017power} with \textbf{low resolution 128$\times$128 based on PLs images only}, (b) dataset \cite{zhang2019detecting} based on \textbf{manually cropping PLs images only}, (c)  \textbf{most of images are not aerial images} \cite{russakovsky2015imagenet}, and (d) our dataset (TTPLA) on TTs and PLs images without manually cropping.}
	\label{fig:publicdataset}
	\vskip-5mm
	\end{figure}
The paper is organized as follows. Prior work is discussed in Section~\ref{sec2}. The properties of the TTPLA dataset are presented in Section~\ref{sec3}. Evaluation of TTPLA is demonstrated in Section~\ref{sec4}. Finally, conclusions are summarized in Section~\ref{sec5}.

\section{Related Work\label{sec2}}
There have been several research papers released recently based on their published or unpublished datasets on TTs, PLs, and insulators.  In this section we will review these datasets, reported in Table~\ref{T:relateddatasets}, with the understanding that there is still a lack of training datasets for TTs and PLs in general, as mentioned in~\cite{sampedro2014supervised,nguyen2019intelligent,candamo2008wire}, due to the difficulty in image collection, especially when UAVs fly close to power~grids.

\subsection{Datasets Based on TTs Images Only}
\label{s2:sub3}
The dataset, introduced in \cite{sampedro2014supervised}, is employed to detect TTs from the aerial images. The dataset consists of 3,200 images in which only 1,600 images contain towers while the rest contains background only. The images size is 64$\times$128 pixels which is scaled down from the original frame sizes (550$\times$480 and 720$\times$576). Objects are labeled by bounding boxes.  In \cite{nguyen2019intelligent}, four datasets are presented. Only one dataset with 28,674 images is annotated with bounding boxes with image size 6,048$\times$4,032 pixels. The other three datasets are binary labeled and built on cropped images with image size 256$\times$256. These datasets are exploited to detect and classify TT damage such as missing top caps, cracks in poles and cross arms, woodpecker damage on poles, and rot damage on cross arms.  The dataset in \cite{hui2018vision} includes 600 aerial images for training and testing with resolution 1,280$\times$720 pixels. Unfortunately, all of these datasets are not available to the public.

\begin{table}[th]%
	\vskip-6mm

	\caption {Related Datasets.}
	\label{T:relateddatasets}\centering %
	\begin{tabular}{l|l|l|l|l|l|l|l}
		\toprule %
		Target  &Dataset  &Public  &Image$\#$(Pos.) &Image Size & Annotation Type & Syn.  & Manual\\\hline
		TTs &\cite{sampedro2014supervised}  & No  & 3,200(1,600) &64$\times$128  & bounding box & No & No\\
        TTs &\cite{nguyen2019intelligent} & No  &28,674 &6,048$\times$4,032 & bounding box  & No & No\\
		 TTs &\cite{hui2018vision}  &No  & 600 &1,280$\times$720 & Binary Mask  & No & No\\\hline
		PLs &\cite{emre2017power} & Yes    & 4,000(2,000) & 128$\times$128 & Binary Classif. & No & No\\
		 PLs &\cite{zhang2019detecting} & Yes   & 573 &540$\times$360  & Binary Mask &No & Yes\\
		PLs &\cite{zhang2019detecting} & Yes   & 287 &540$\times$360  & Binary Mask &No & Yes\\
	     PLs &\cite{saurav2019power} &No   & 3,568 &5,12$\times$512  & Binary Mask &No & Yes\\
	    PLs	&\cite{nguyen2019ls}& No  &718,000 &- & Binary Mask &Yes & No \\
		 PLs &\cite{madaan2017wire}& No   &67,000 &480$\times$640  & Binary Mask &Yes & No \\ \hline
		 Both & \cite{russakovsky2015imagenet} & Yes  & 1,290 &Various  & Class Label & No & No \\\hline \hline
		 \textbf{Both} &\textbf{TTPLA} & \textbf{Yes}   & \textbf{1,100} & \textbf{3,840$\times$2,160}  & \textbf{Instance Seg.} & \textbf{No} & \textbf{No}\\\hline
	\end{tabular}
	\vskip-8mm
\end{table}

\subsection{Datasets Based on PLs Images Only}
\label{s2:sub1}
Two datasets are presented in~\cite{emre2017power} on PL images with video resolutions 576$\times$325 pixels for infrared and 1,920$\times$1,080 pixels for visible light, respectively.  The first dataset is built on image-level class labels while the second dataset consists of binary labels at the pixel-level.  Only 2,000 images among the 4,000 images under visible light in the datasets include PLs while the rest does not.  The image size is scaled down to 128$\times$128 pixels from the original video sizes as shown in Fig. \ref{fig:publicdataset} (a).  The image-level class labels are exploited for binary classification training. In the work~\cite{yetgin2018power}, the dataset in~\cite{emre2017power} is employed by two CNN-based PL recognition methods to identify whether the images contain PLs without the consideration of localization. 
The work in~\cite{zhang2018automatic} also relies on the datasets in~\cite{emre2017power} by resizing the images to 224$\times$224 pixels. A CNN model is used as a binary classifier to identify whether PLs are present in images. Ground truth of the PL dataset consists of 400 infrared and 400 visible light images with the resolution of 512$\times$512 pixels. 
\smallskip
\\ 
\textbf{Datasets with Manual Cropping.}
Two public datasets on PL images are presented in \cite{zhang2019detecting}, including urban scene and mountain scene captured by UAVs. The Urban scene dataset consists of 453 images for training and 120 images for testing, while the mountain scene dataset consists of 237 and 50 images for training and testing, respectively.  The original image size is 3,000$\times$4,000. However, the images are manually cropped to meaningful regions of 540$\times$360 pixels to get close scenes for PLs as shown in Fig. \ref{fig:publicdataset} (b). Pixel-level annotation is used to label the cropped images in both datasets. VGG16 architecture~\cite{simonyan2014very} is modified based on richer convolutional features~\cite{liu2017richer} to evaluate both datasets.  
The dataset in~\cite{saurav2019power} includes 530 PL images captured by UAV with the resolution of 5,472$\times$3,078 pixels.  These images are manually cropped and divided into non-overlapped patches with the size of 512$\times$512 pixels.  Then all patches that do not contain any PLs are removed.  The total number of images is 3,568 with the size of 512$\times$512 pixels. Nested U-Net architectures are evaluated on this dataset.

In general, manually cropping images may not be practical for real-time UAV operations.  UAVs can fly from any directions, which means that TTs and PLs can appear in any region of the images. Manually manipulated images cannot reflect the noisy backgrounds that UAVs may face in real life.  Alternatively, automatic image cropping and zooming can be applied in lane detection problems to get the region of interest, because of bird-view imaging~\cite{zou2019robust}.
\smallskip
\\
\textbf{Synthetic Datasets.} 
There are two datasets using synthetic PLs.  In the first dataset~\cite{nguyen2019ls},  synthetic images of power lines are rendered using the physically based rendering approach to generate the training dataset.  Synthetic PLs are randomly superimposed on 718k high dynamic range images collected from the internet. In addition, data augmentation techniques are used to increase the amount of training data.  
In the second dataset~\cite{madaan2017wire}, the synthetic wires from a raytracing engine are superimposed on 67k images. These images are extracted from 154 flight videos available on the internet with image resolution 480$\times$640~\cite{madaan2017wire}.  Both datasets are not publicly available. 
\subsection{Datasets Based on Both TTs and PLs Images}
\label{s2:sub2}
ImageNet~\cite{russakovsky2015imagenet} is regarded as one of the largest datasets for object detection, which includes 1,290 annotated images with labels on TTs and PLs.  Most of them are not aerial images as shown in Fig. \ref{fig:publicdataset} (c) and there is no top and side view for TTs and PLs. Imaging from the ground provides most of images simple backgrounds such as sky and white clouds, which may be impractical in our scenario since we focus on UAV applications.
\section{TTPLA Dataset Properties}
\label{sec3}
Building large-scale aerial datasets is a complicated task, including recording data, extracting and selecting images, and establishing the required annotation policy. In the following subsections we will introduce the procedures in our dataset preparation as well as the properties of the TTPLA dataset.

\subsection{Aerial Videos Collection and Images Preparation}

\label{s3:sub1}
Recorded videos are collected by a UAV, Parrot-ANAFI, in two different states in USA to guarantee the varieties of the scenes.  The locations are randomly selected without any intentions and treatments to avoid noisy background.  The UAV contains 4k HDR camera and up to 2.8$\times$ lossless zoom.  Zooming is exploited when collecting the video data, in order to guarantee high-resolution of the objects, such as PLs, without manual cropping.  The TTPLA dataset is extracted from a set of totally 80 videos. All aerial videos have the resolution of $3,840\times2,160$ with 30 fps.
\subsection{TTPLA Dataset Policy}
\label{s3:sub2}

\begin{figure}[ht]
	\centering
		\vskip-4mm
	\includegraphics[width=0.9\textwidth]{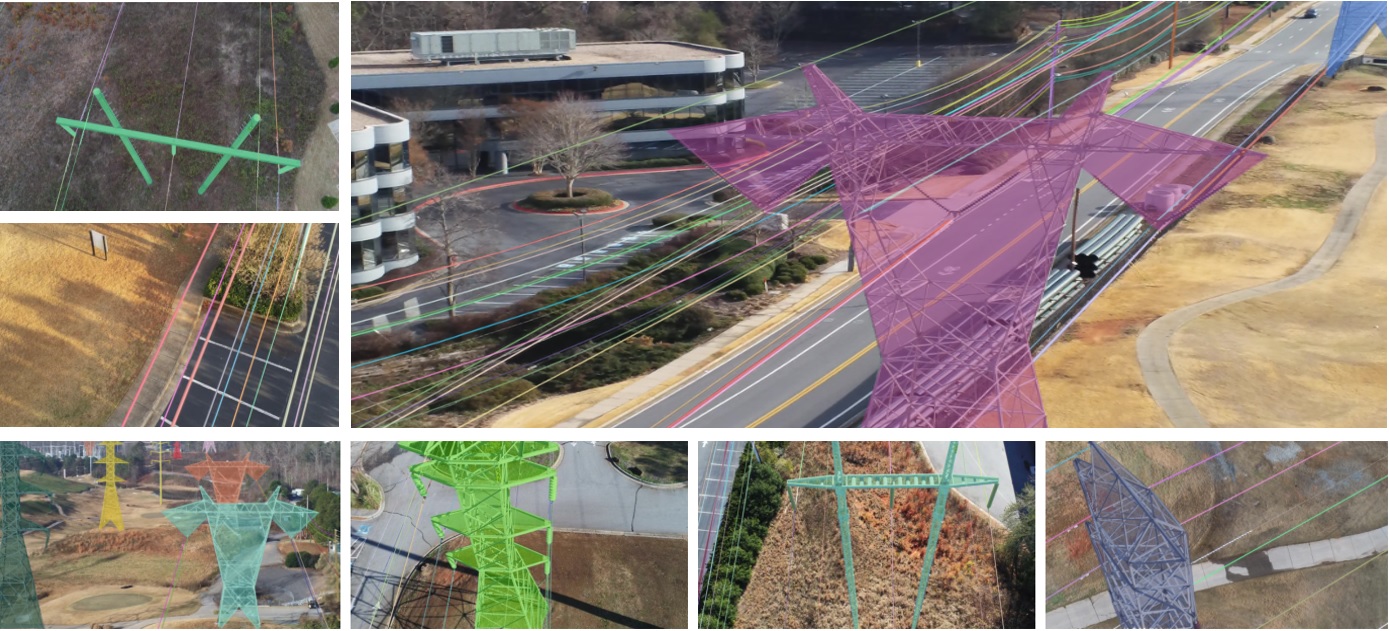}
	\caption{Samples of annotated images in TTPLA.}
	\label{fig:mask}
\vskip-5mm
\end{figure}

Creating a dataset of TTs and PLs needs policies to deal with the diversity of objects during aerial imaging.  For instance, towers are built by different materials (e.g., tubular steel, wood, and concrete) with different structures (e.g., single pole, H-frame, horizontal structure, delta structure and guyed structure) and different insulators~\cite{fang1999transmission}.  Meanwhile, given the shape characteristics of PLs (thin and long), different backgrounds and illumination levels play important roles in PL detection.  With these considerations, we introduce the following policy in data collection and annotation. 
\\ \\
\textbf{Recording Characteristics.}
\label{s3:subsub1} 
The aerial images in TTPLA dataset are extracted from videos taken by UAVs.  The following discussions focus on four important aspects when recording these videos. 
\begin{itemize}
	\vskip-10mm
\item \textit{View angles} are essential in data collection, specially when the shape of the object varies a lot from different view angles.  In TTPLA, all TTs are photographed from different angles such as front view, top view, and side view.  This policy, designed specifically for TTs, guarantees that the deep learning models can detect TTs from any angles.  It provides the freedom to UAVs to fly along any directions without worrying about the detection accuracy.  Various views for different TTs are demonstrated in Fig.~\ref{fig:towerShap}. 
\begin{figure}
	\centering
	\begin{tabular}{c}
	\footnotesize $T_{1}$
	\includegraphics[scale=0.028]{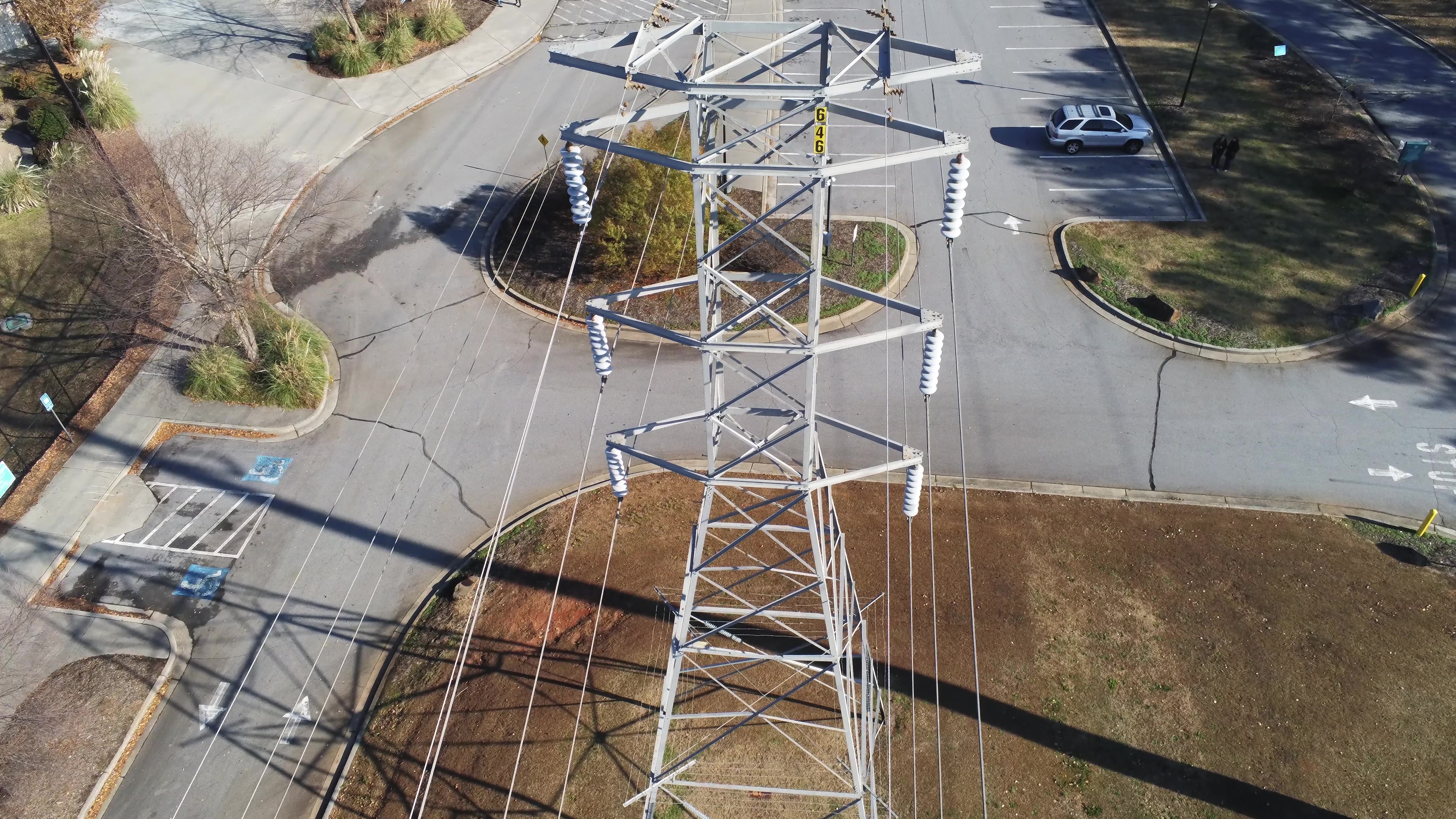}
	\includegraphics[scale=0.028]{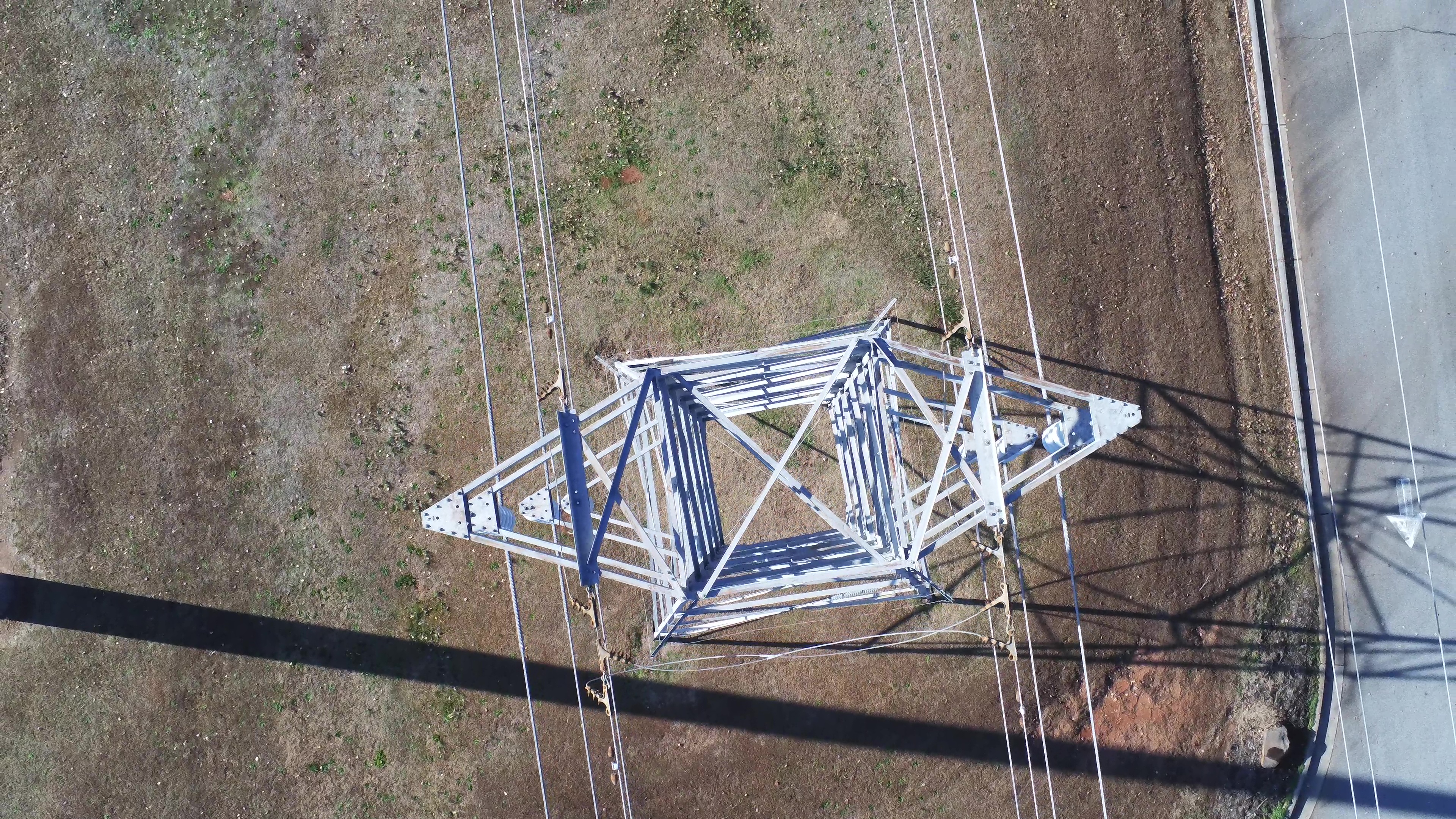}
	\includegraphics[scale=0.028]{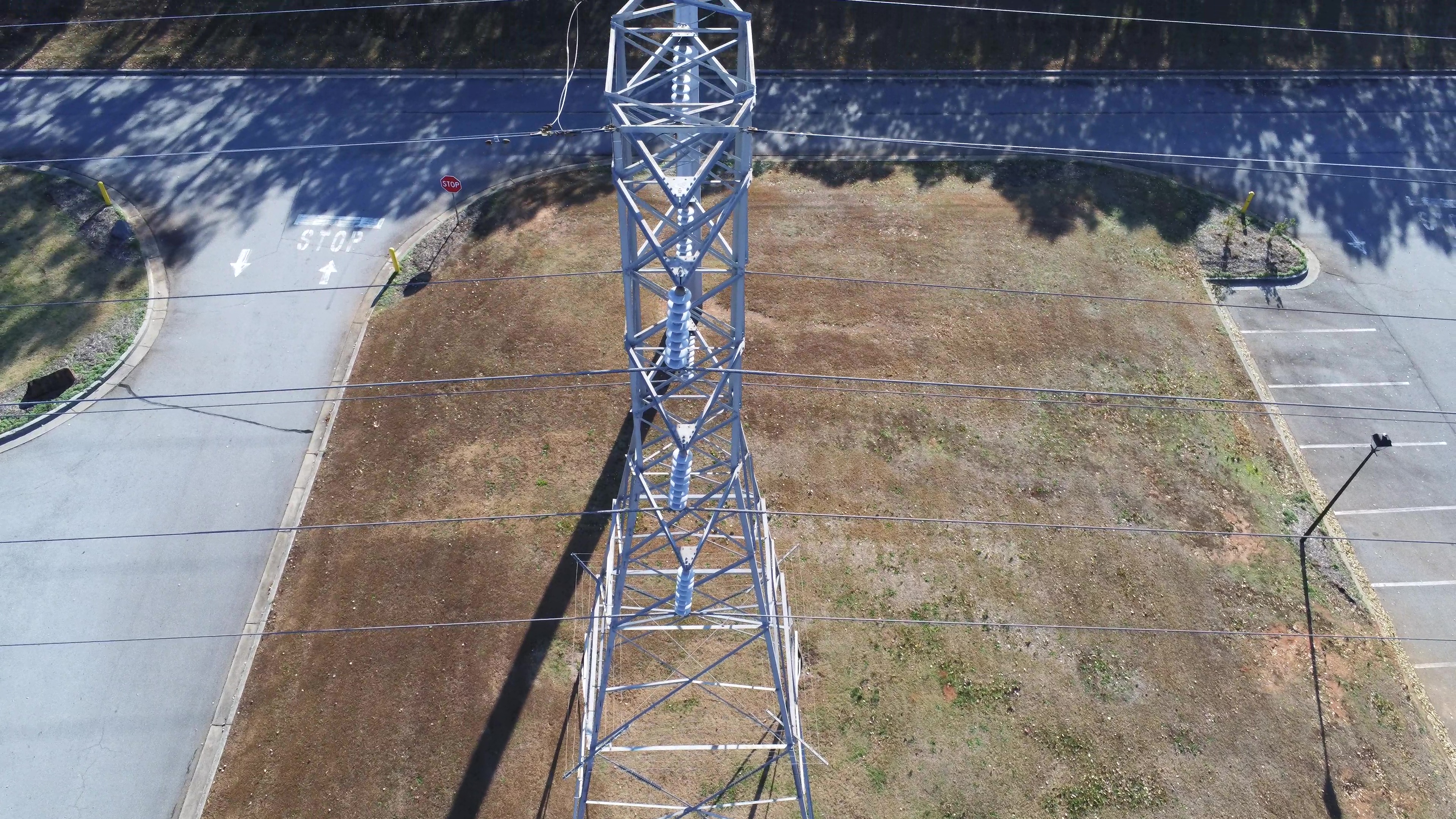}\\
	\footnotesize $T_{2}$
	\includegraphics[scale=0.028]{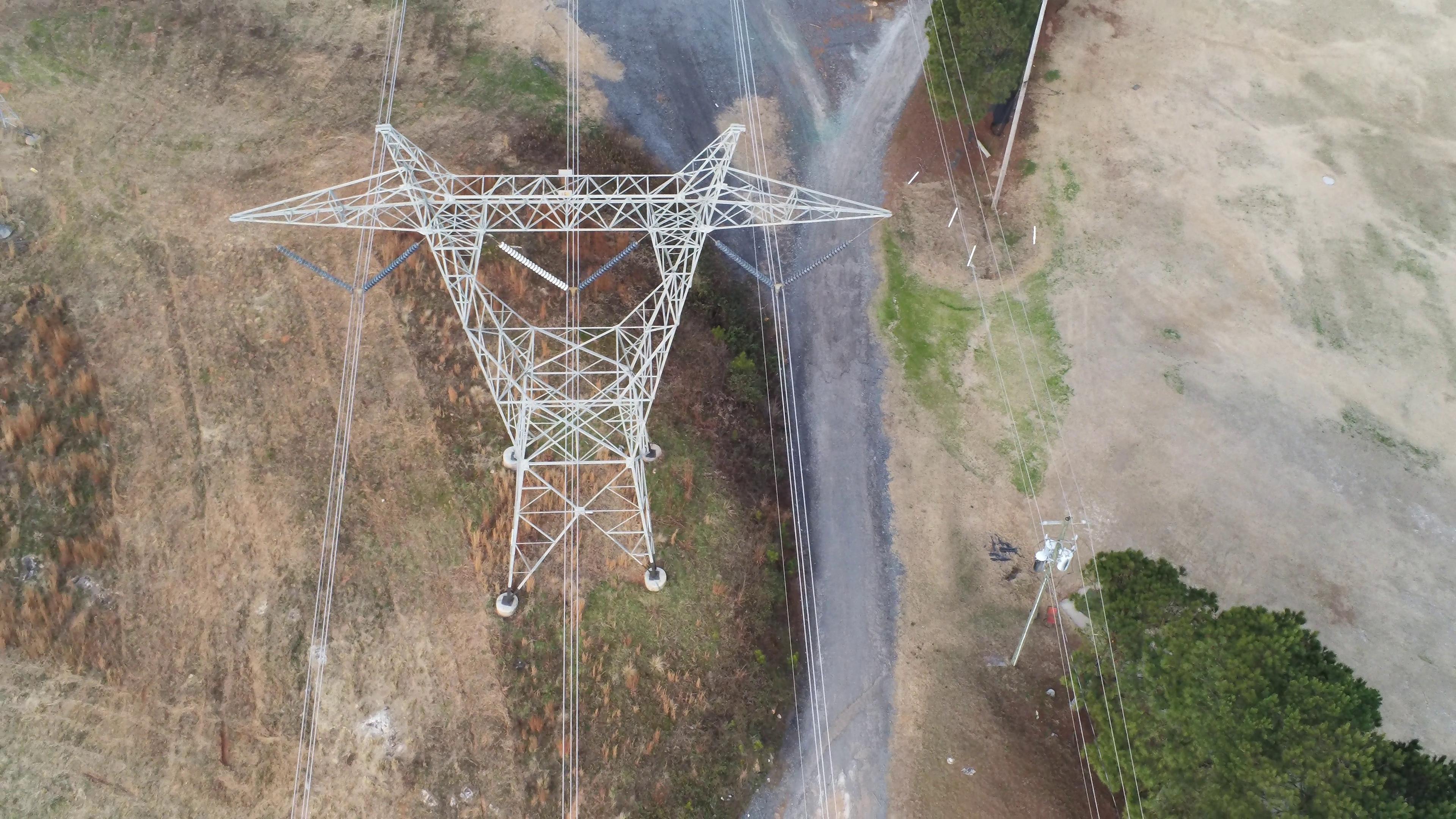}
	\includegraphics[scale=0.028]{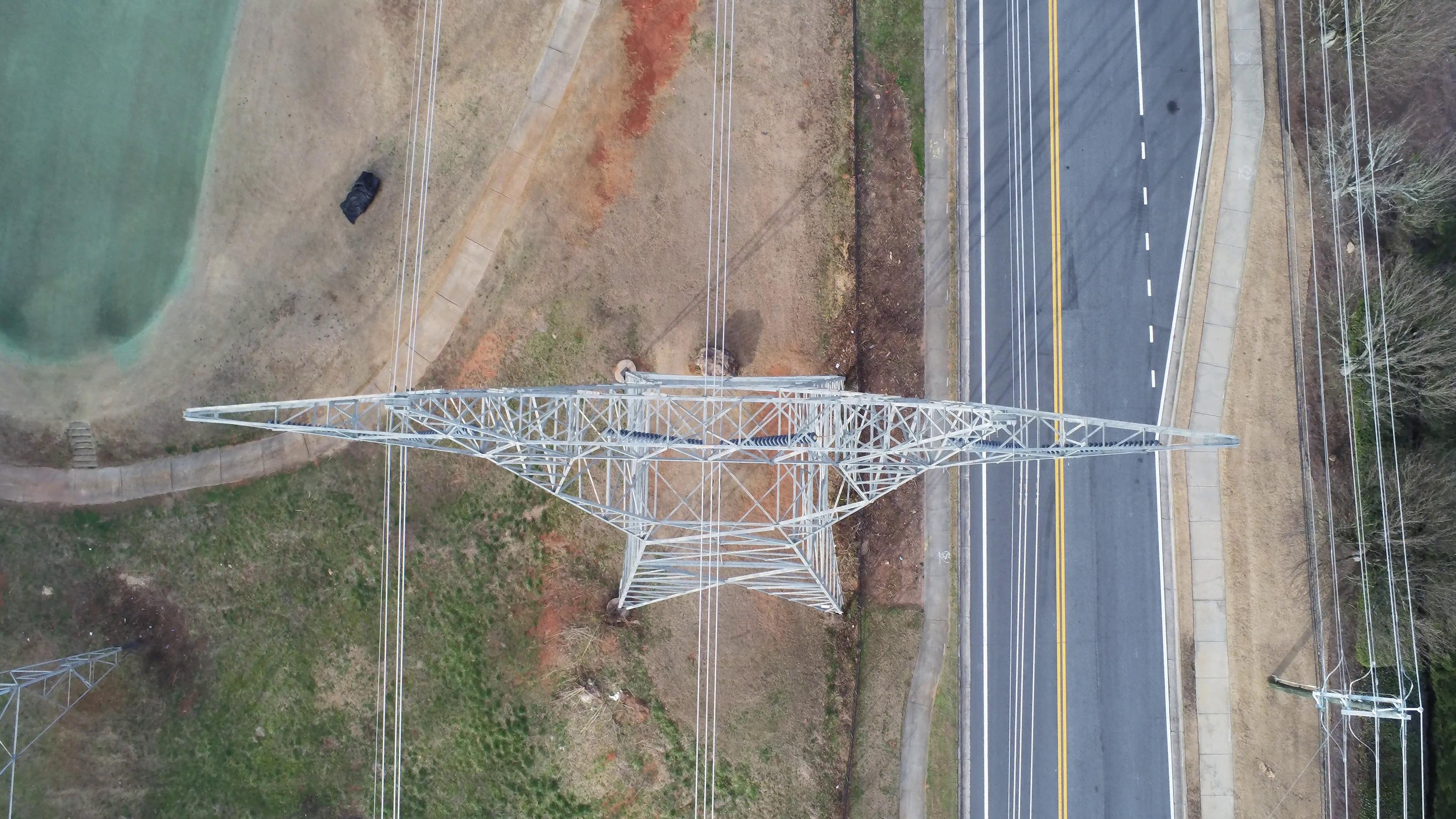}
	\includegraphics[scale=0.028]{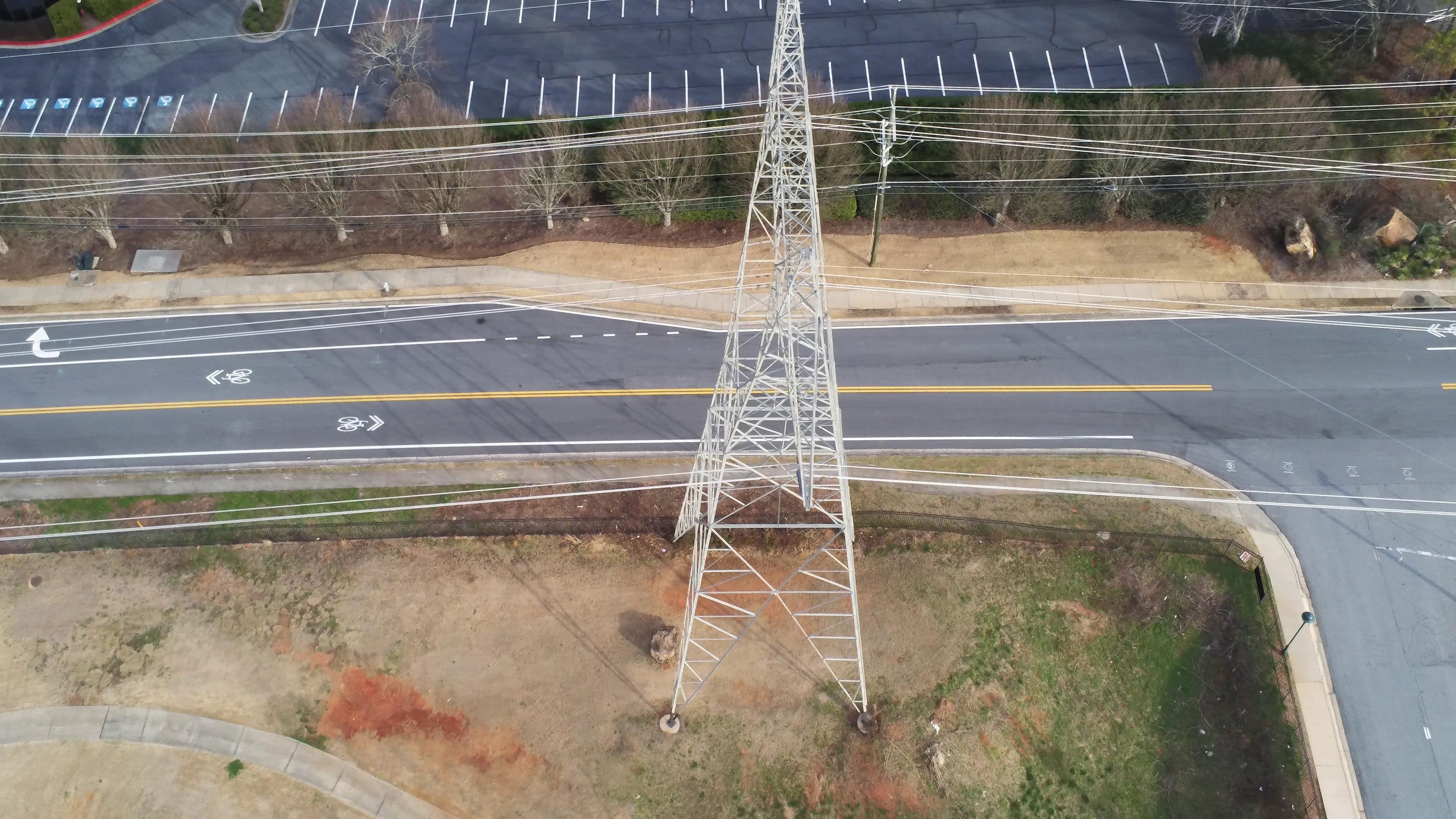}\\
		\footnotesize $T_{3}$
	\includegraphics[scale=0.168]{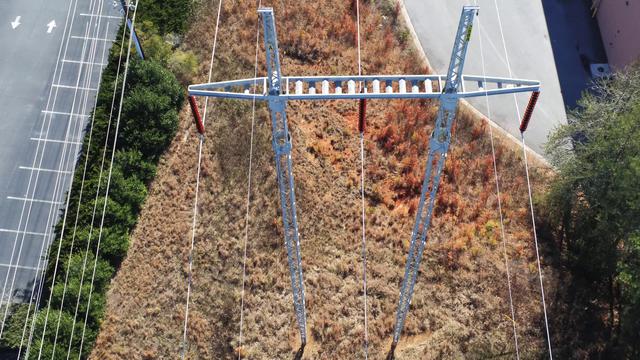}
	\includegraphics[scale=0.168]{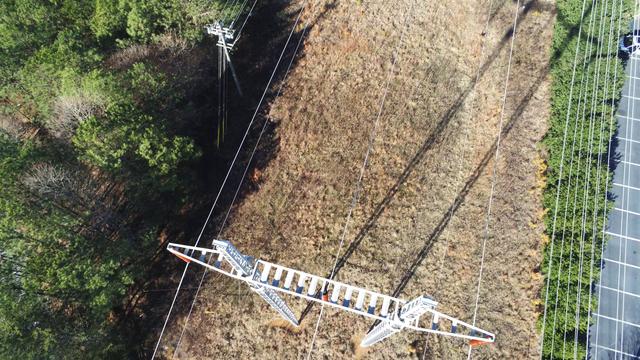}
	\includegraphics[scale=0.168]{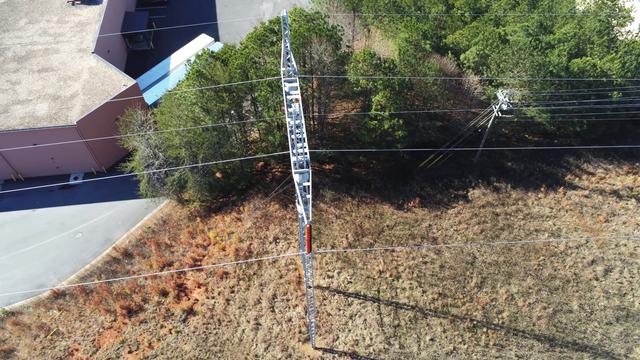}\\
		\footnotesize $T_{4}$
	\includegraphics[scale=0.028]{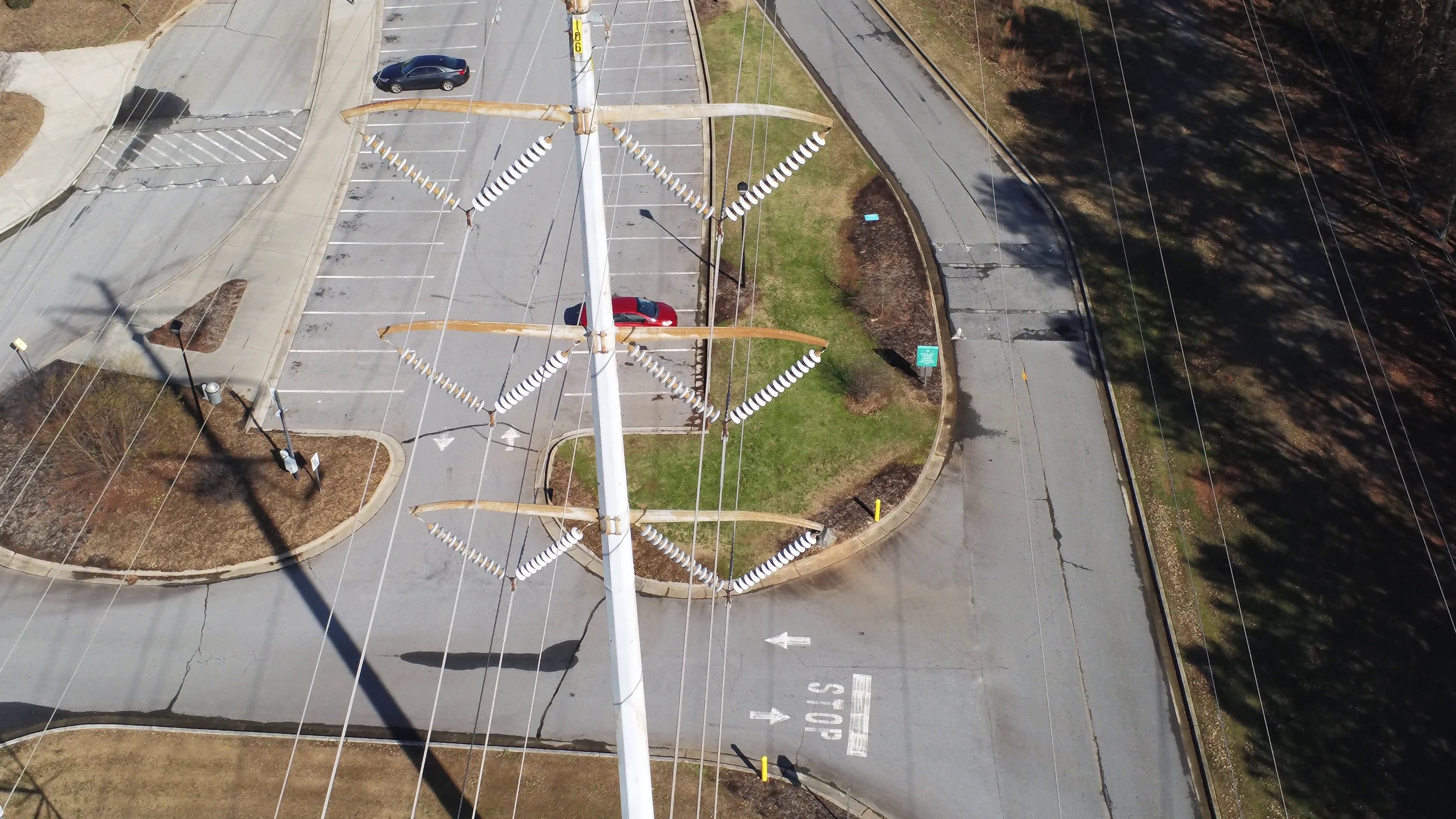}
	\includegraphics[scale=0.028]{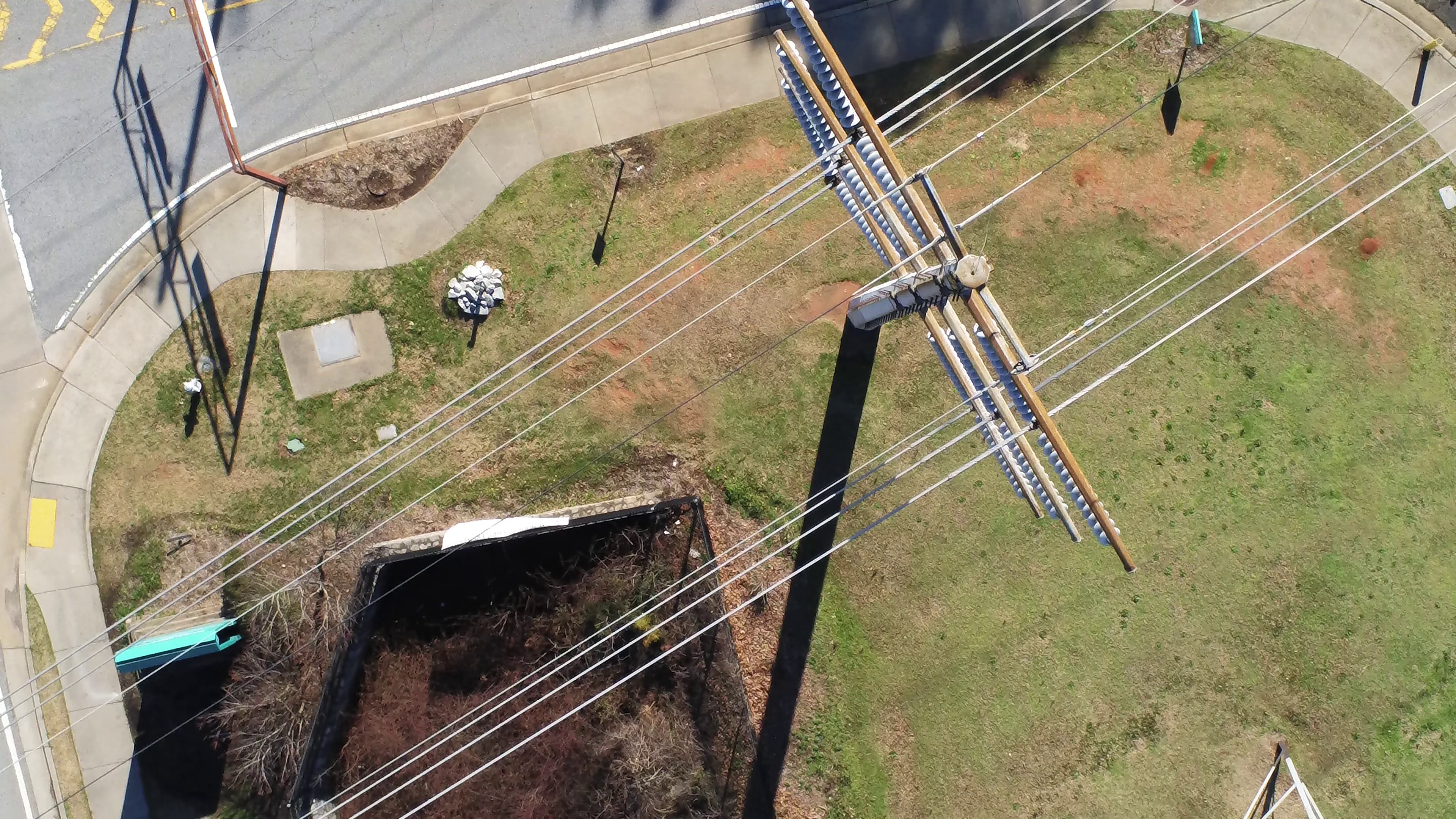}
	\includegraphics[scale=0.028]{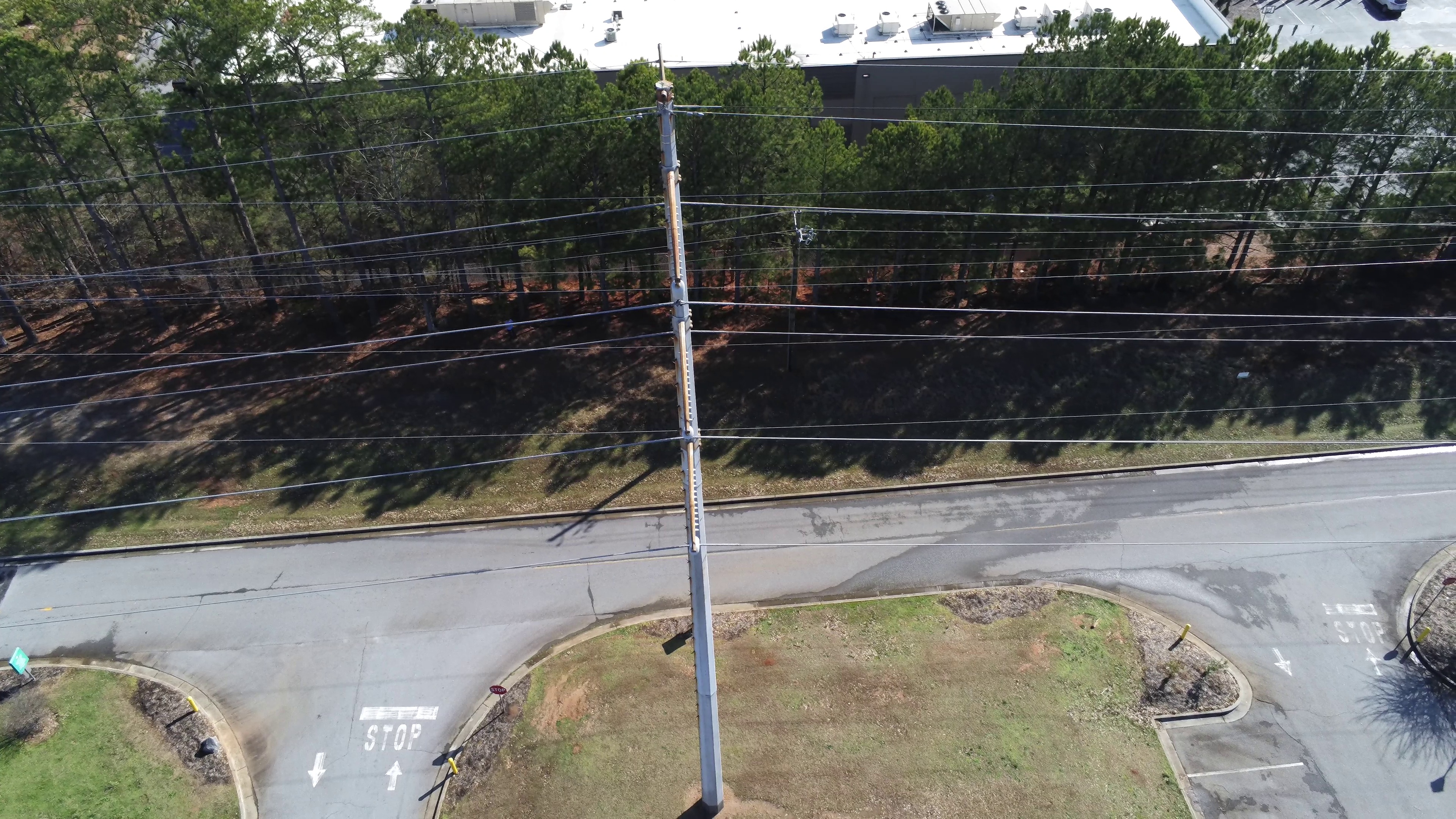}\\
		\footnotesize $T_{5}$
	\includegraphics[scale=0.028]{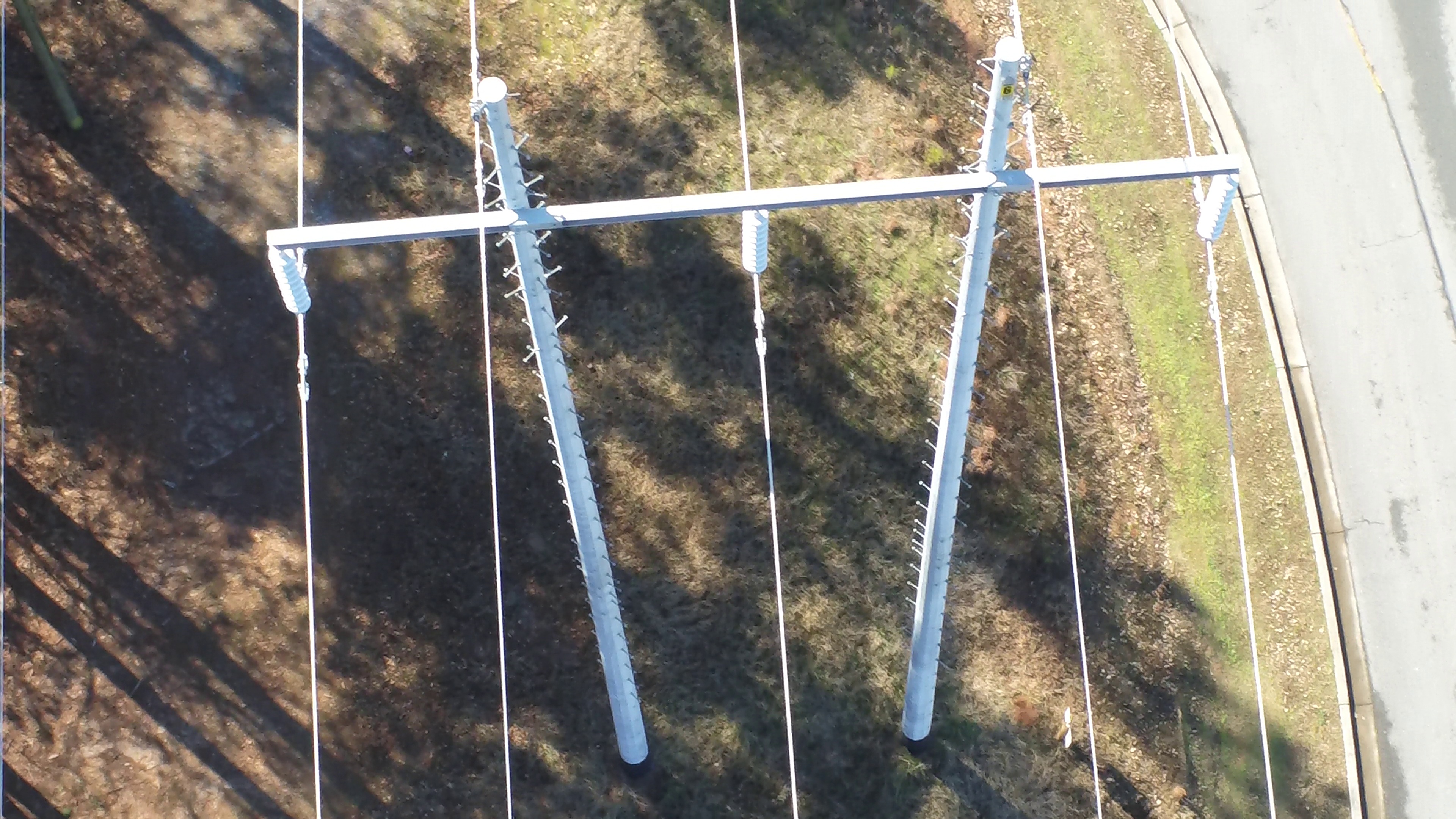}
	\includegraphics[scale=0.028]{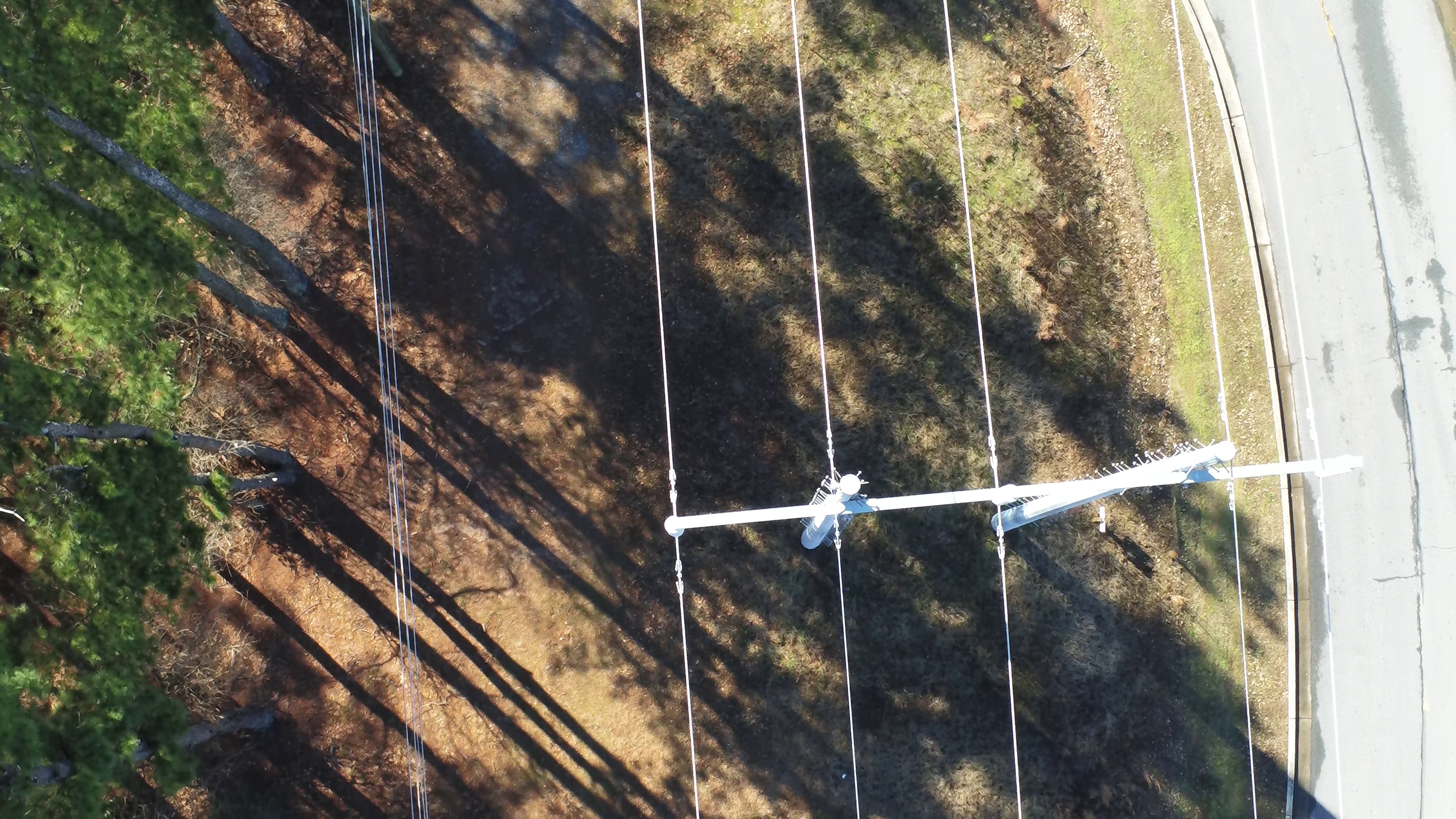}
	\includegraphics[scale=0.028]{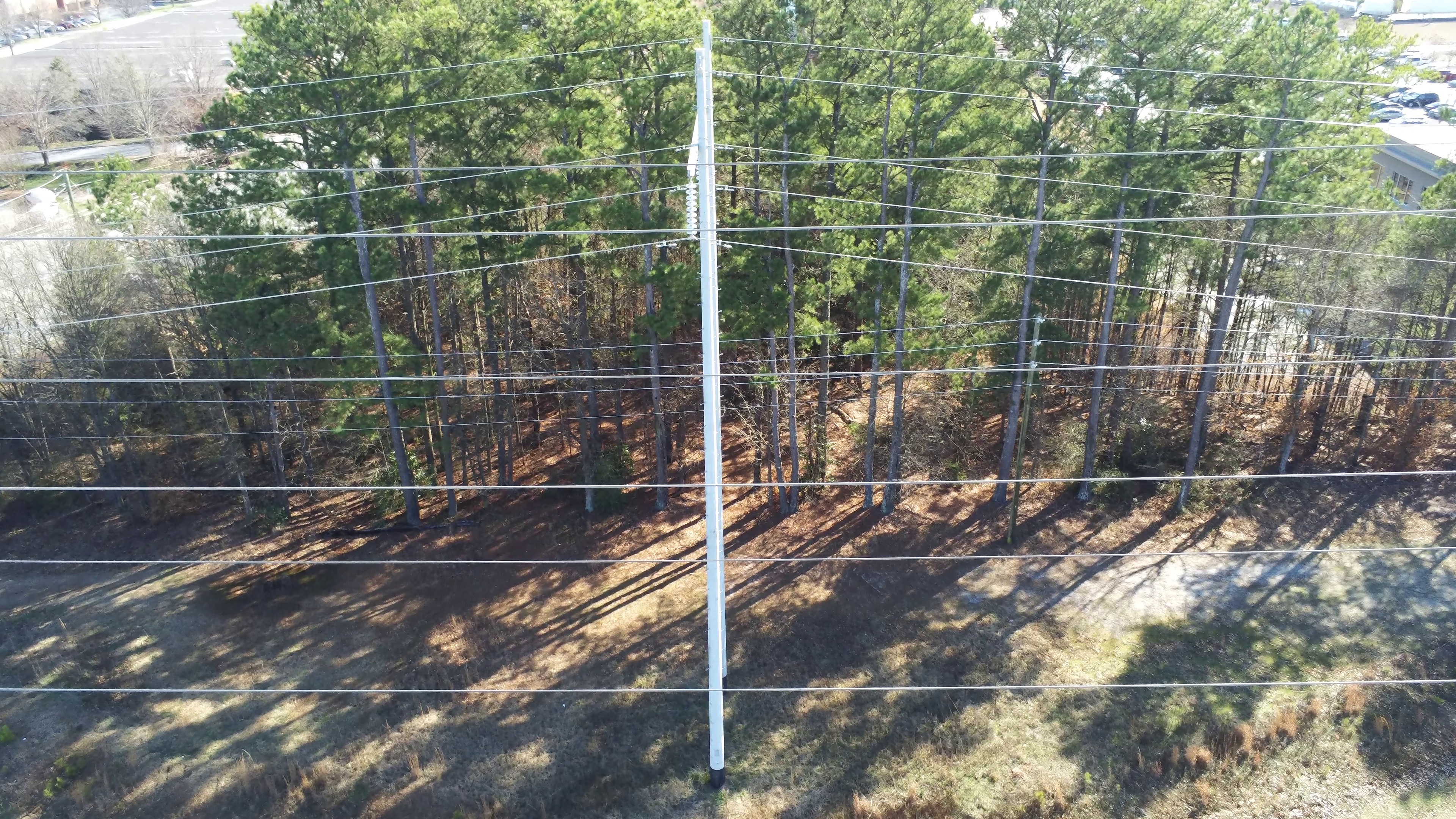}\\
		\footnotesize $T_{6}$
	\includegraphics[scale=0.028]{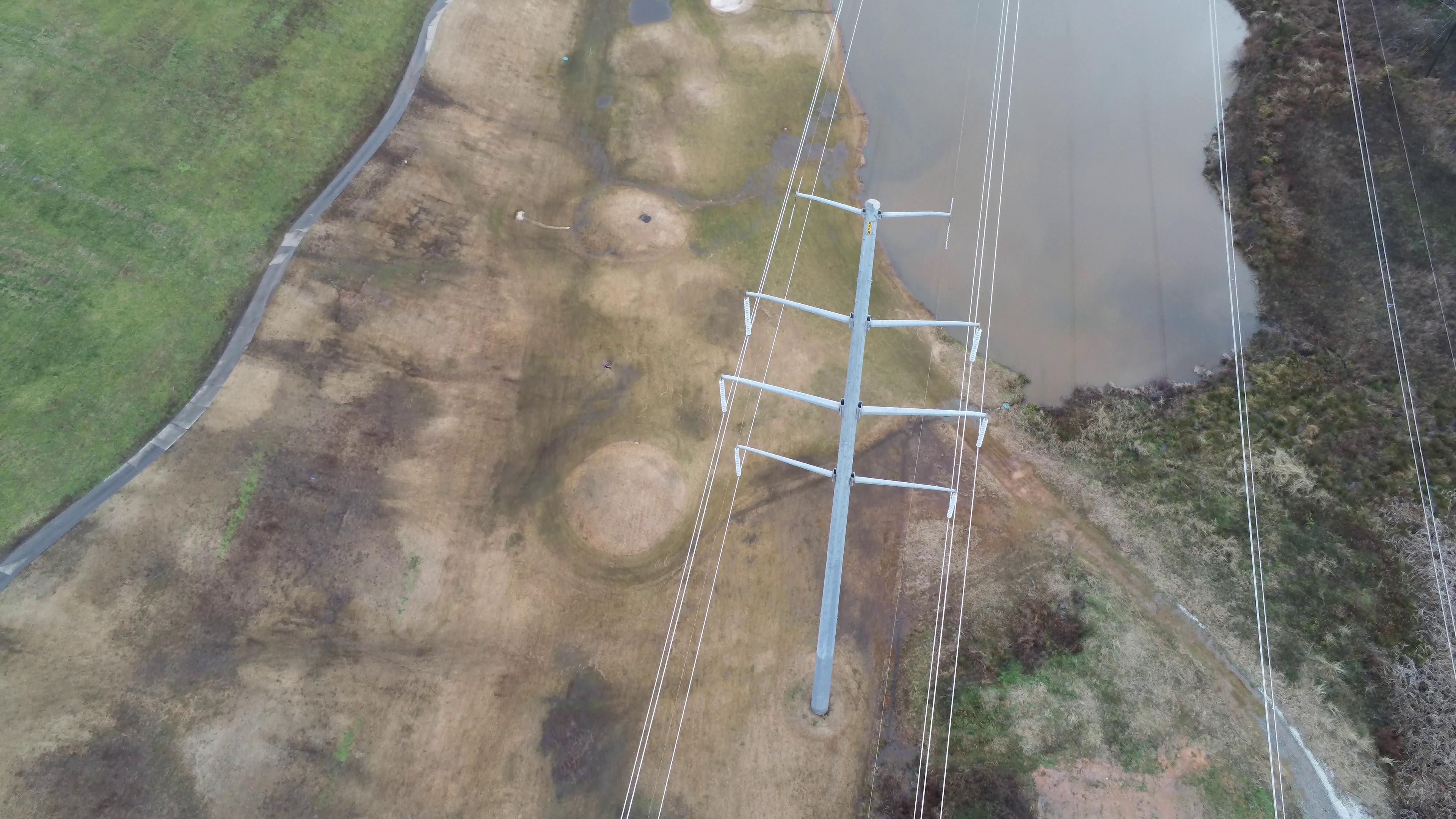}
	\includegraphics[scale=0.028]{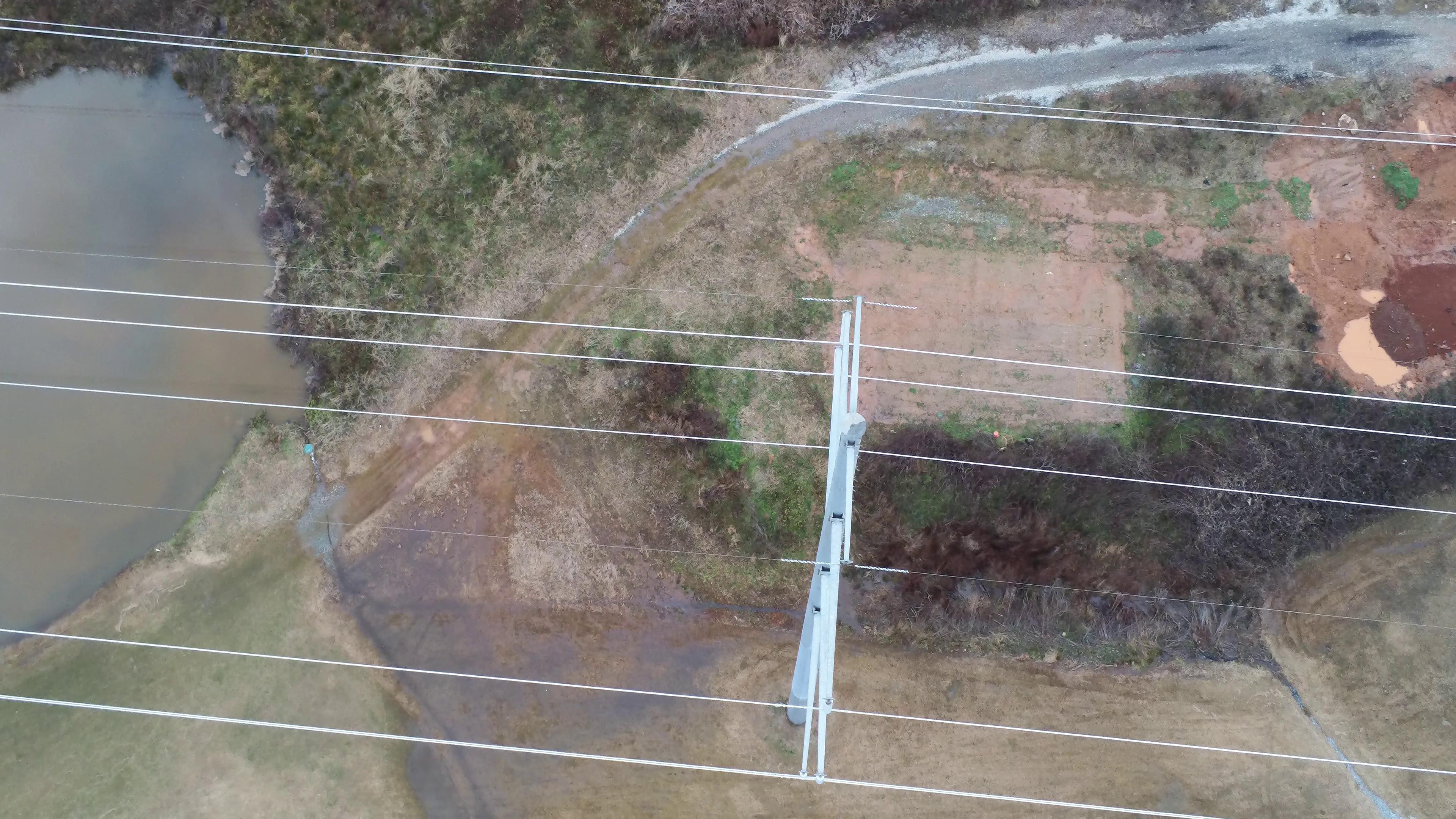}
	\includegraphics[scale=0.028]{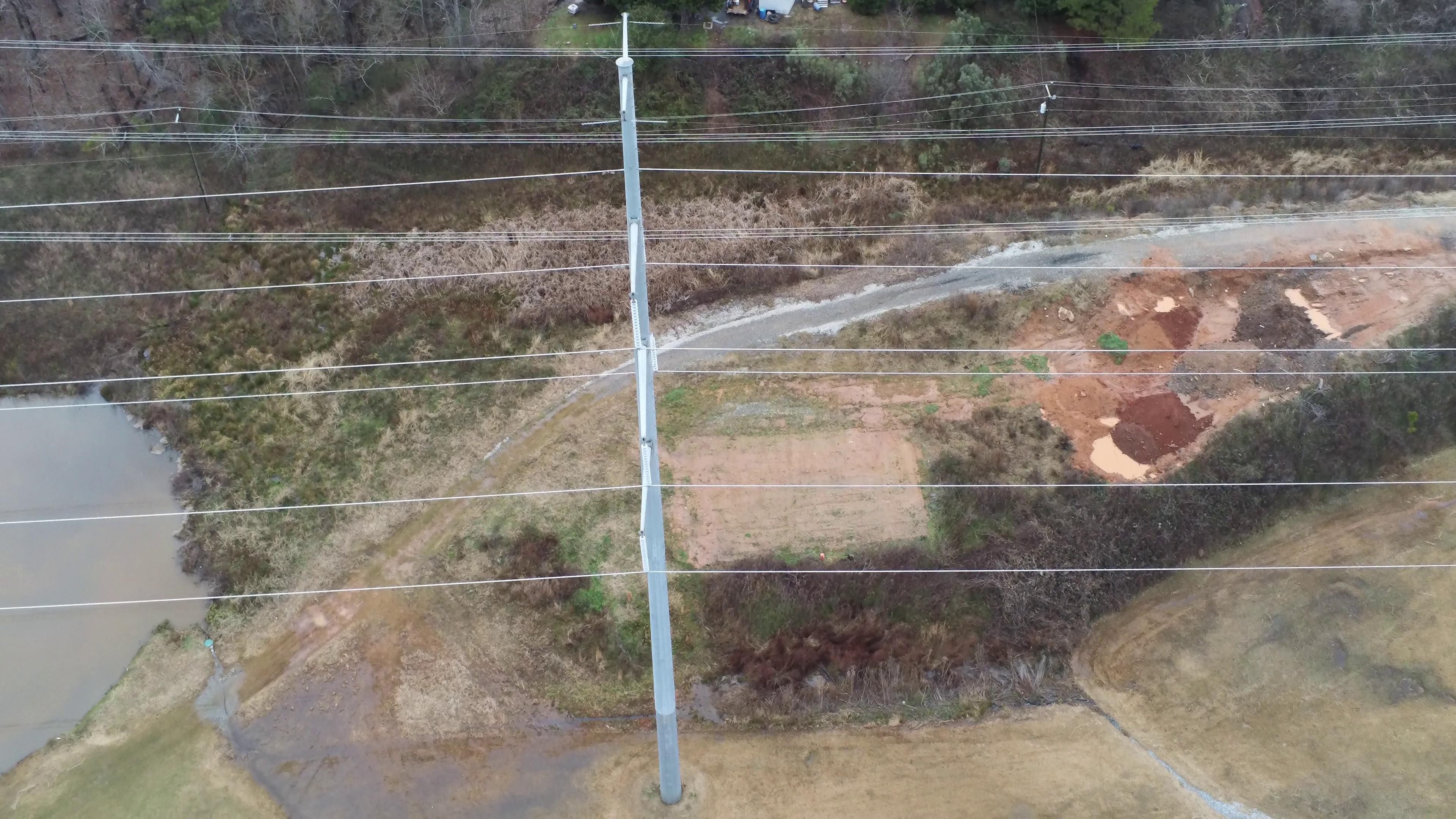}\\
		\footnotesize $T_{7}$
	\includegraphics[scale=0.168]{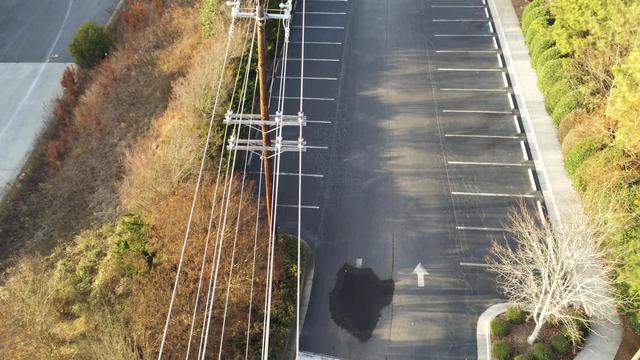}
	\includegraphics[scale=0.168]{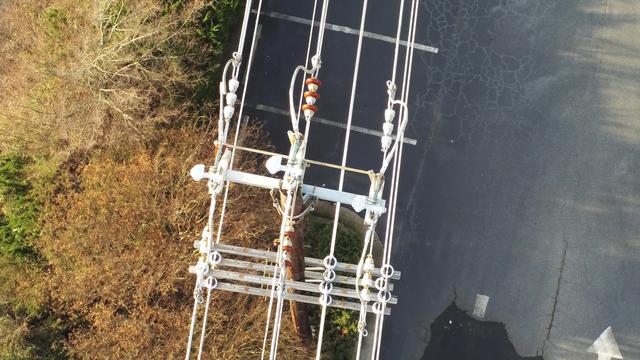}
	\includegraphics[scale=0.168]{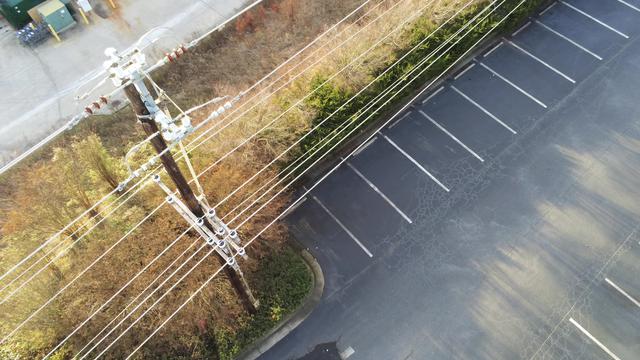}\\
		\footnotesize $T_{8}$
	\includegraphics[scale=0.028]{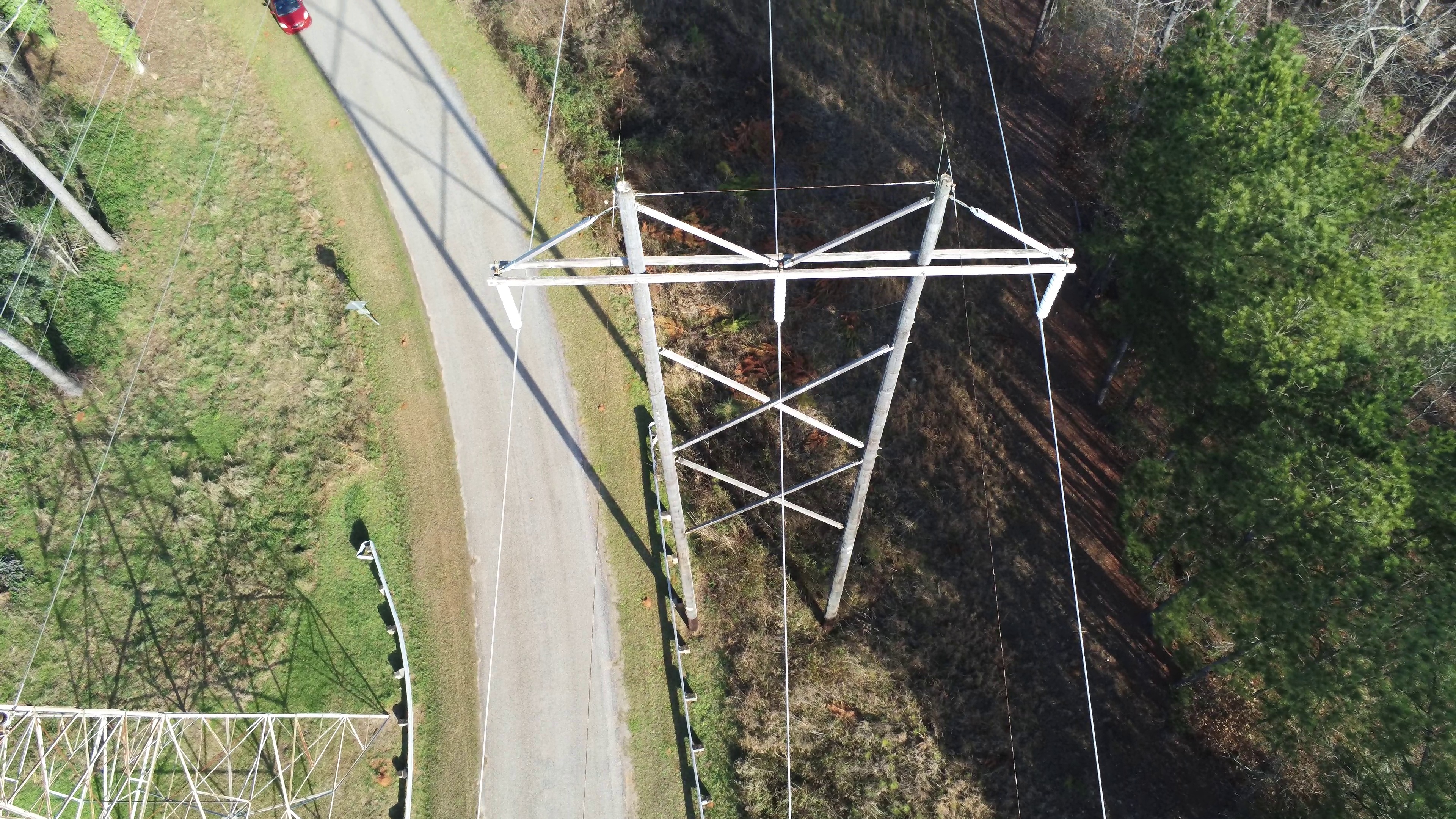}
	\includegraphics[scale=0.028]{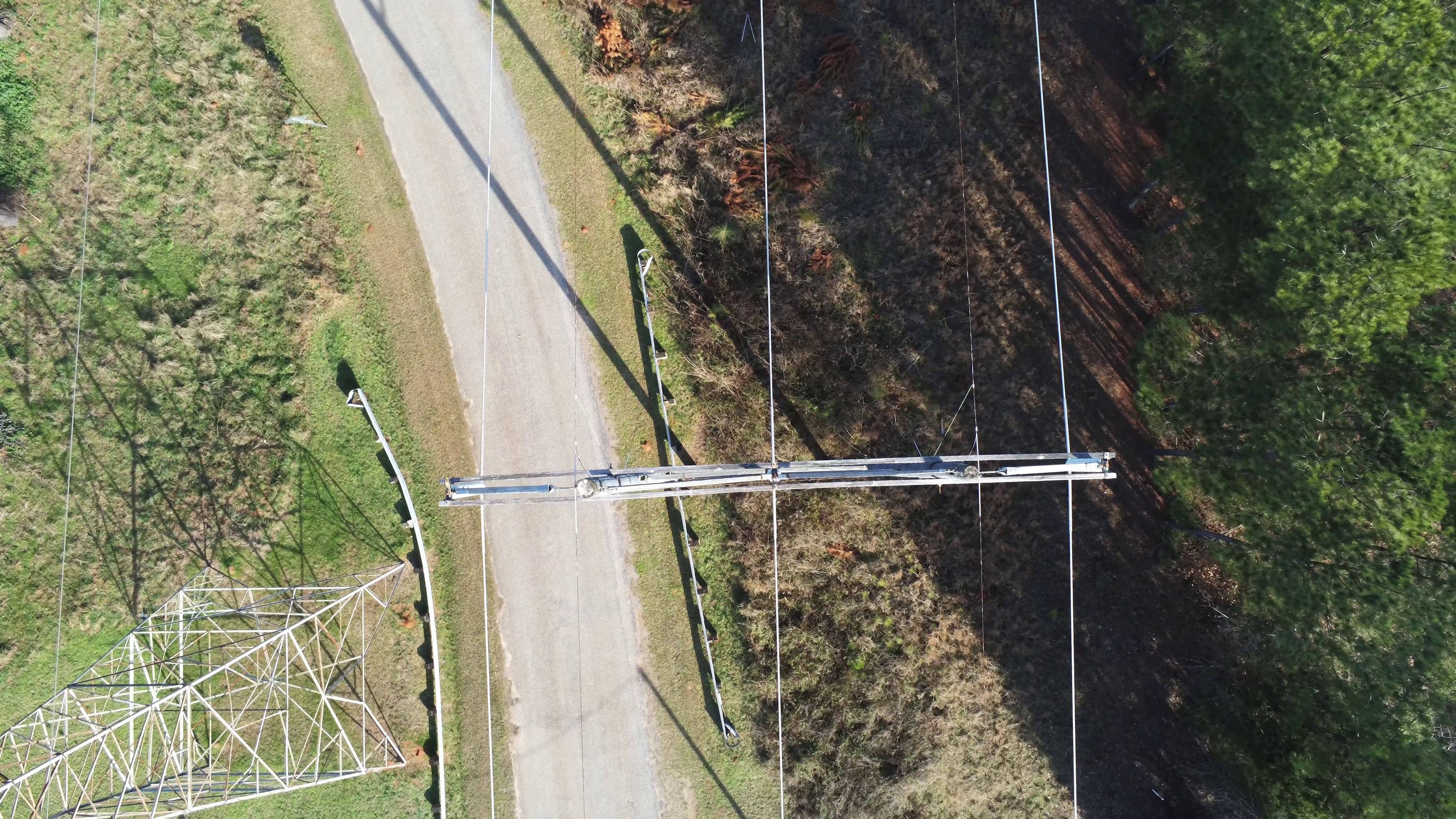}
	\includegraphics[scale=0.028]{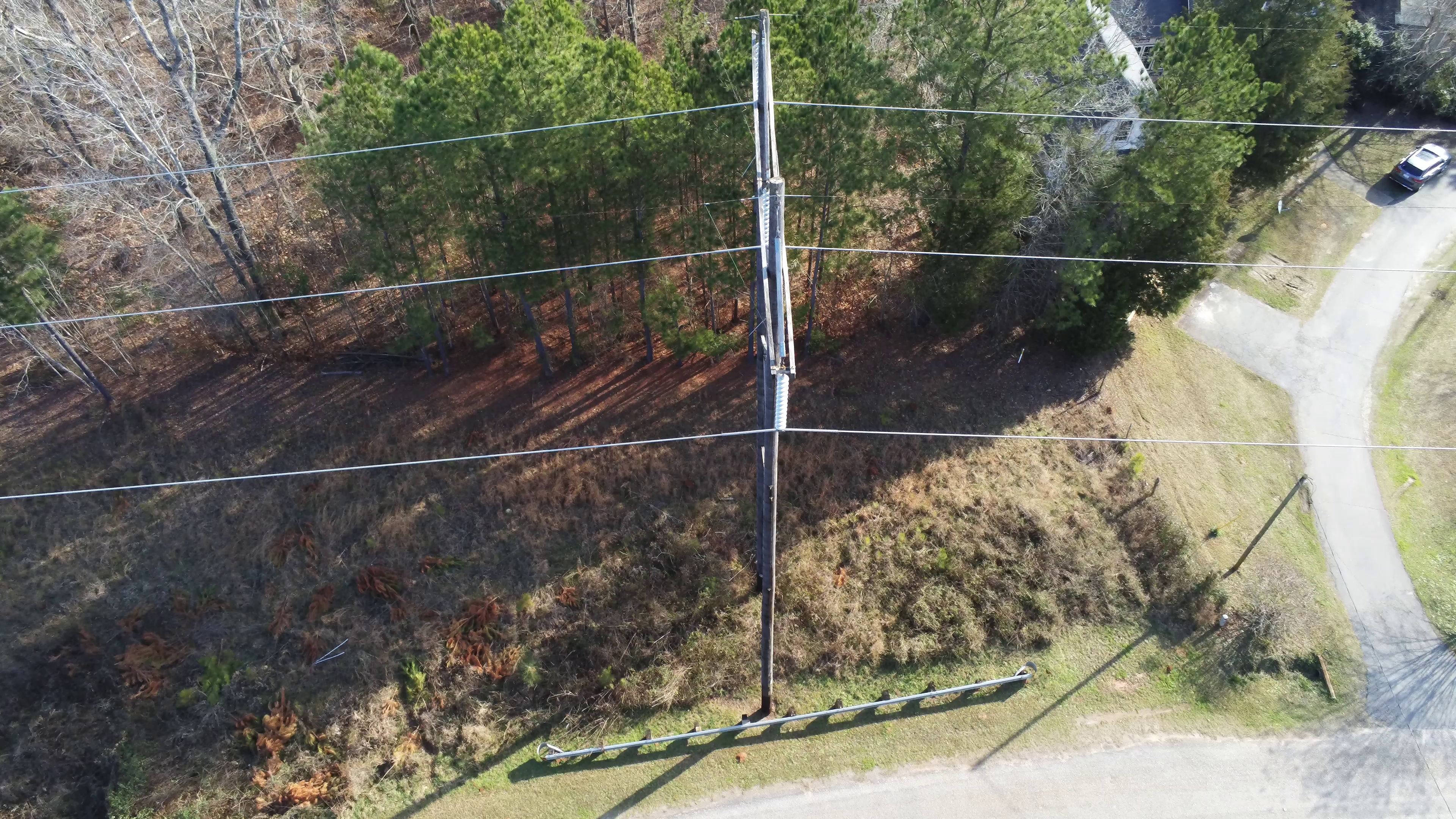}
	\end{tabular}
	\caption{Different types of TTs in TTPLA. Front view, top view and side view are ordered from the left to the right for each TT shape.}
	\label{fig:towerShap}

\end{figure}
\item The images are taken by randomly varying the \textit{zooming level} together with the motion of the camera. Different zooming levels are explored in TTPLA to capture accurate features of PLs, especially with noisy backgrounds as shown in Fig.~\ref{fig:powerline}.
\item The videos are recorded at different time during a day under different weather conditions.
\item \textit{Backgrounds} are important to accurately detect PLs.  From the UAV's viewpoint, the backgrounds of most PLs in images are noisy. TTPLA consists of plentiful PLs images with noisy backgrounds, which make extracting PLs a challenging task due to ``thin and long'' features of PLs~\cite{li2019future}.  Moreover, the color of PLs can be very close to that of the background (e.g., building, plants, road, lane line) and sometimes PLs may be hidden behind the trees.  We include all these cases in TTPLA.
\end{itemize}
\textbf{Preparing Images.}
It is not an easy task to select appropriate frames into TTPLA from a large set of videos.  We consider the following aspects to avoid duplicate images. To ensure that the images are not duplicated, the videos are renamed with unique IDs and each video is separately archived.  The lengths of the recorded videos are between 1 min. to 5 min, which imply 1,800 to 9,000 images per video, given 30fps. These images are then sampled once every 15 frames before manual inspection, which means 2 images per second.  If necessary, we can extract 3 images per second to augment the dataset~\cite{zou2019robust}.  The next step is manual inspection of the selected images, given the possibility that sometimes PLs are not clear in images due to day light or complex backgrounds. Another reason of having manual inspection is to make sure that the whole views of TTs are included to keep our recording policy consistent.  In addition, manual inspection removes all redundant images from the dataset.  Finally, the selected images are renamed by the related video ID, followed by the related frame~number.
\begin{figure}[ht]
	\centering
	\includegraphics[scale=0.021]{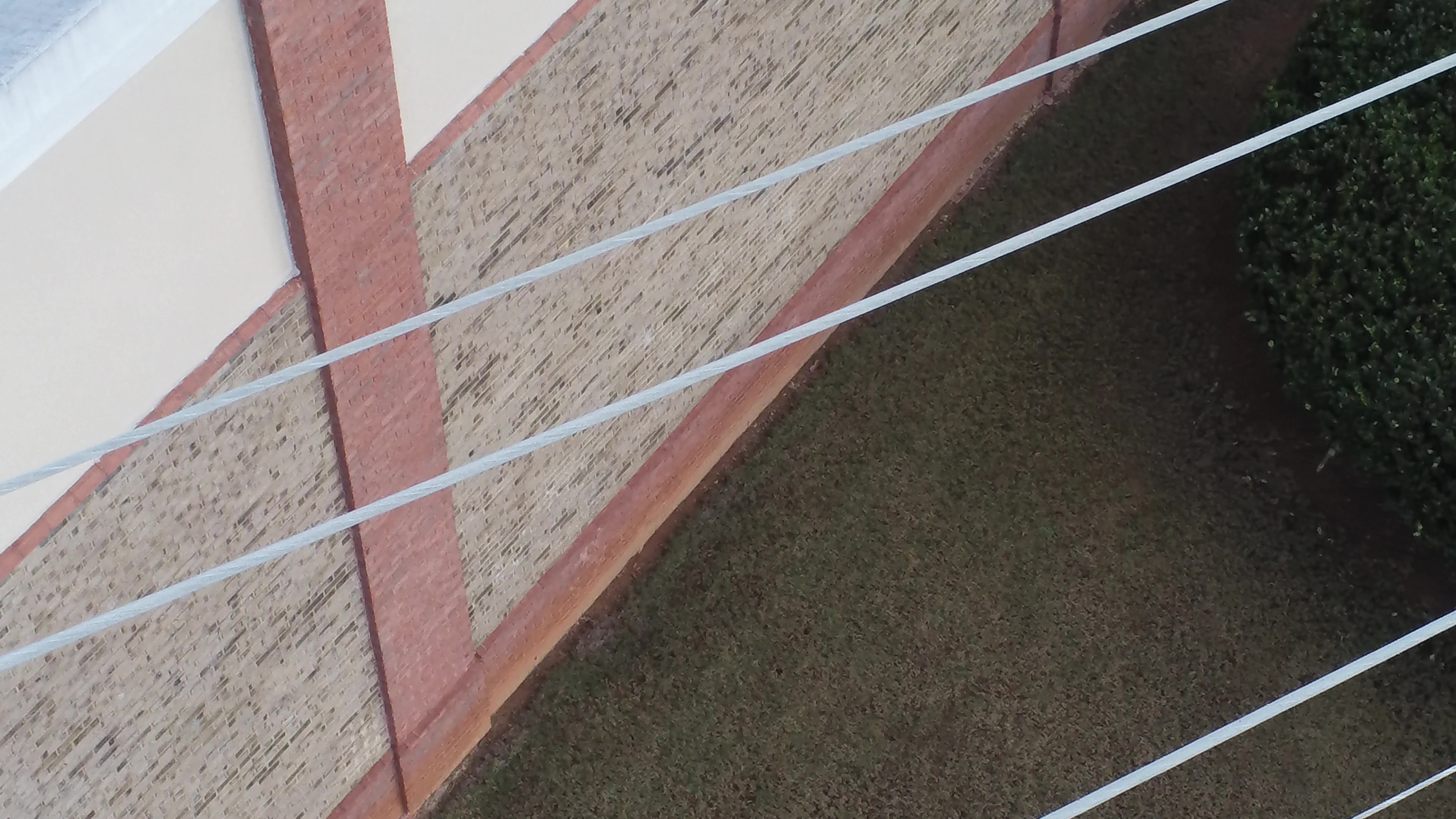}
	\includegraphics[scale=0.125]{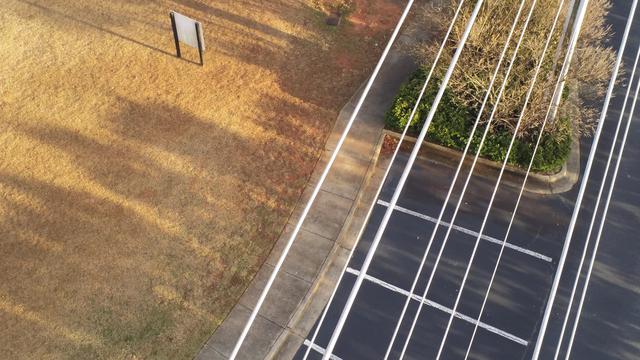}
	\includegraphics[scale=0.125]{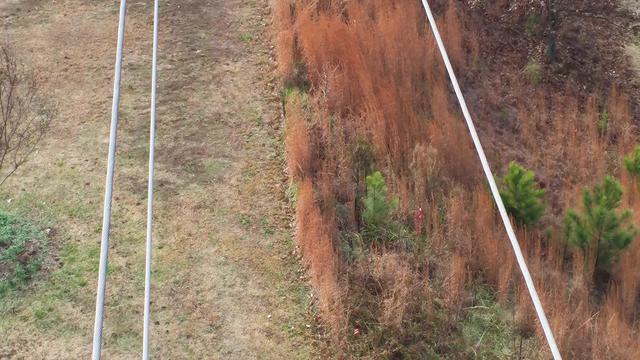}
	\includegraphics[scale=0.021]{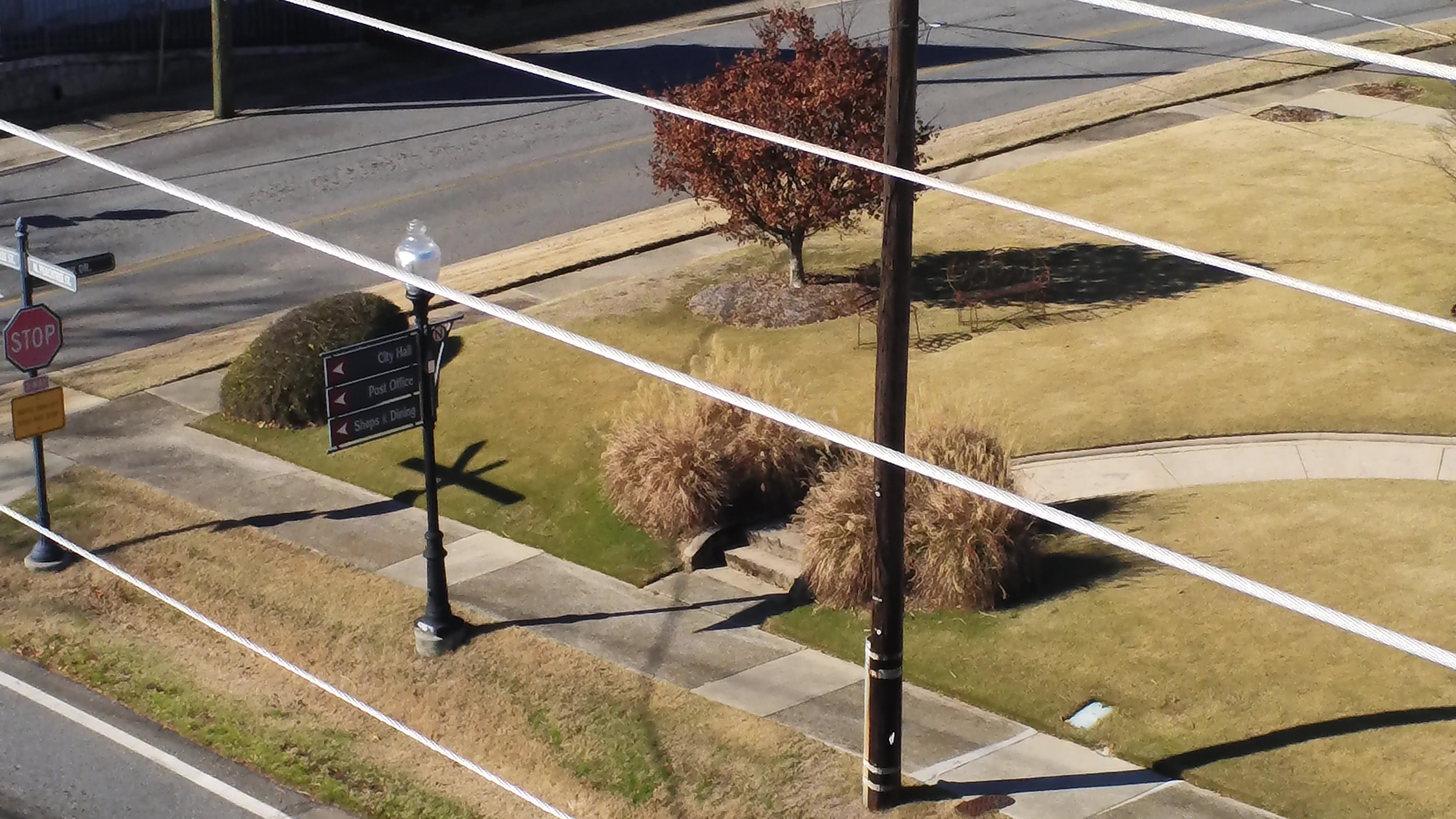} \\ 	\vskip+1mm
	\includegraphics[scale=0.021]{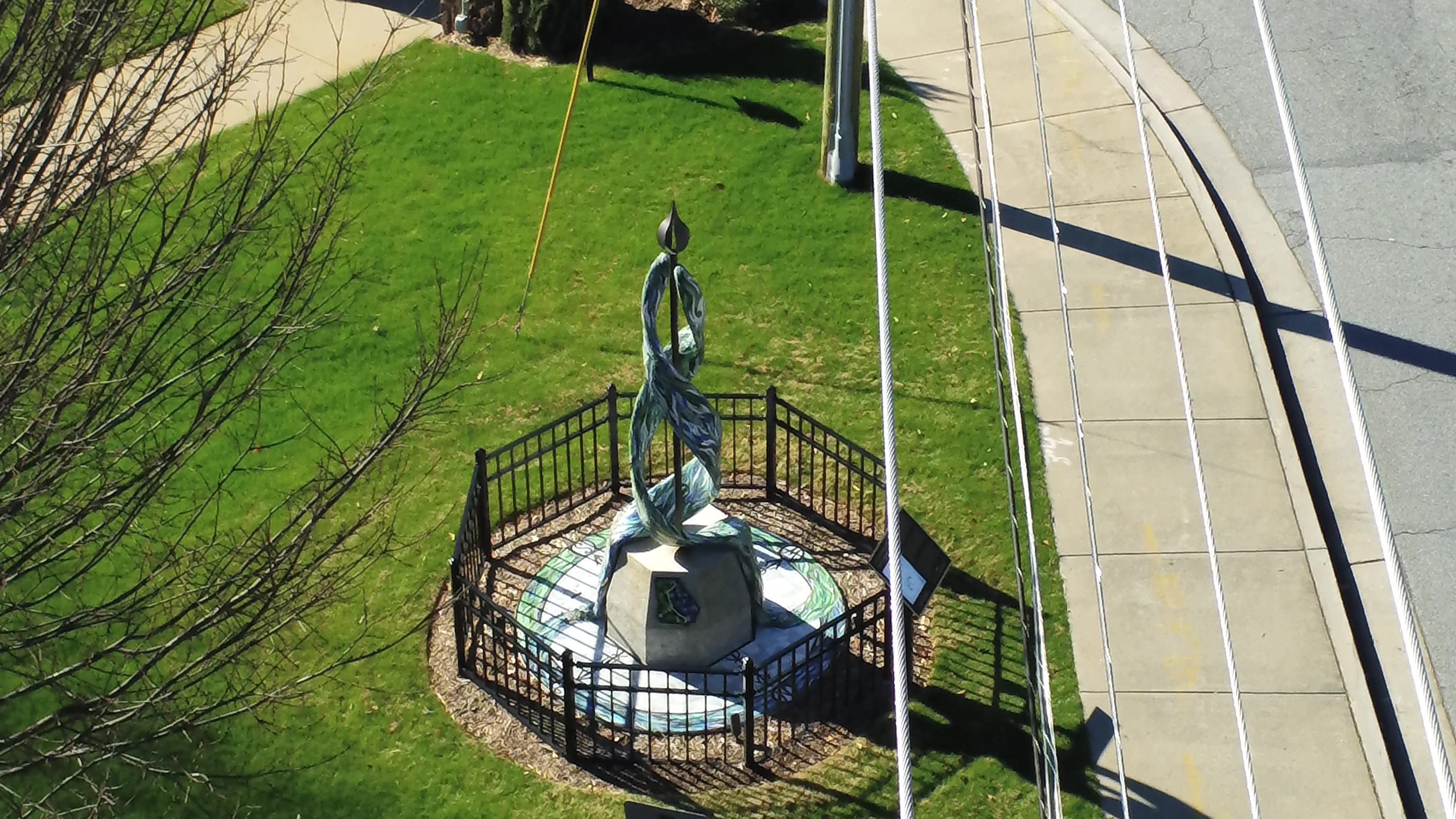}
	\includegraphics[scale=0.021]{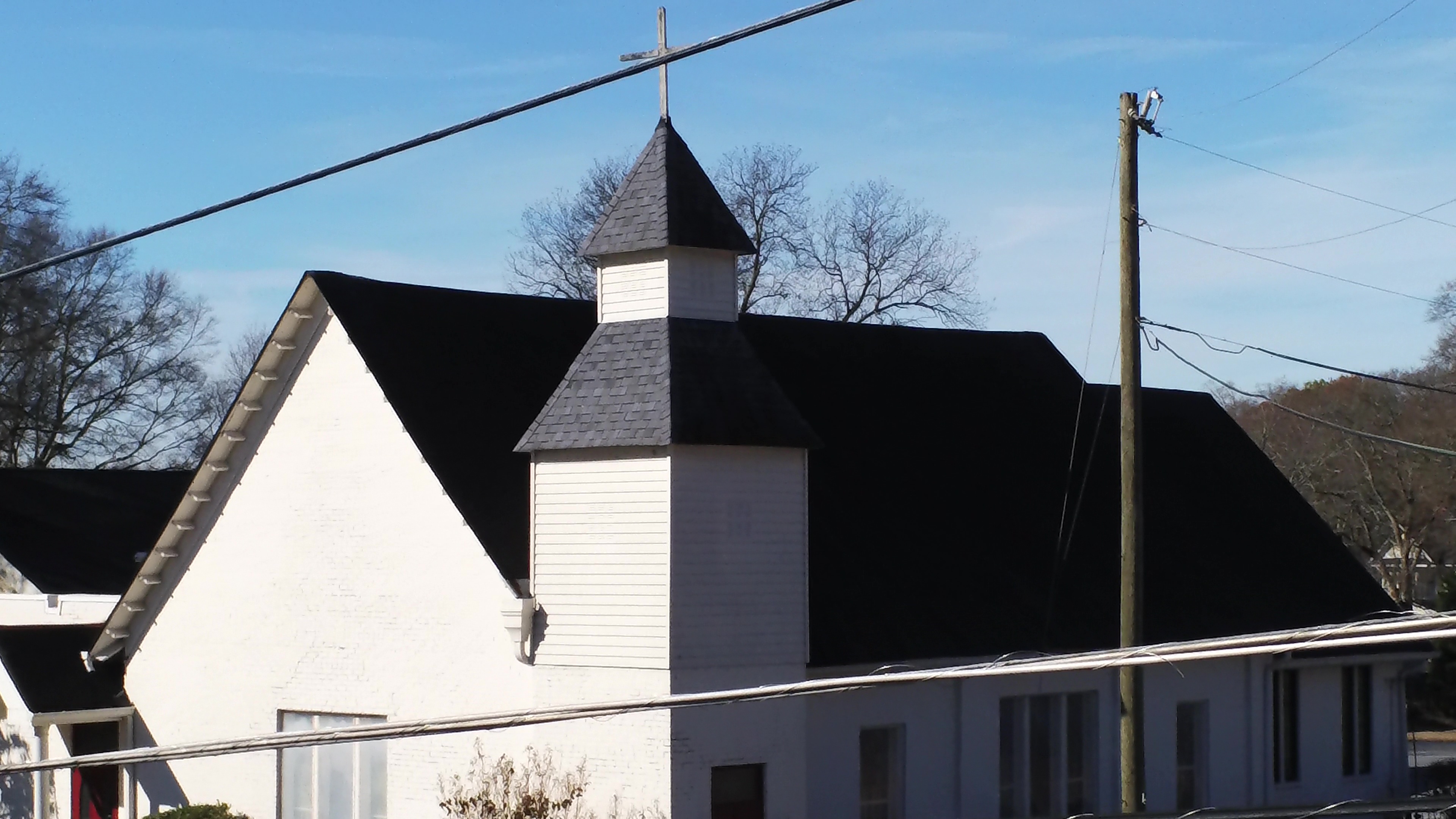}
	\includegraphics[scale=0.125]{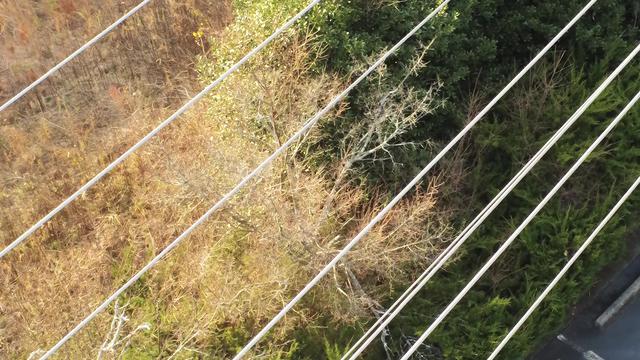}
	\includegraphics[scale=0.021]{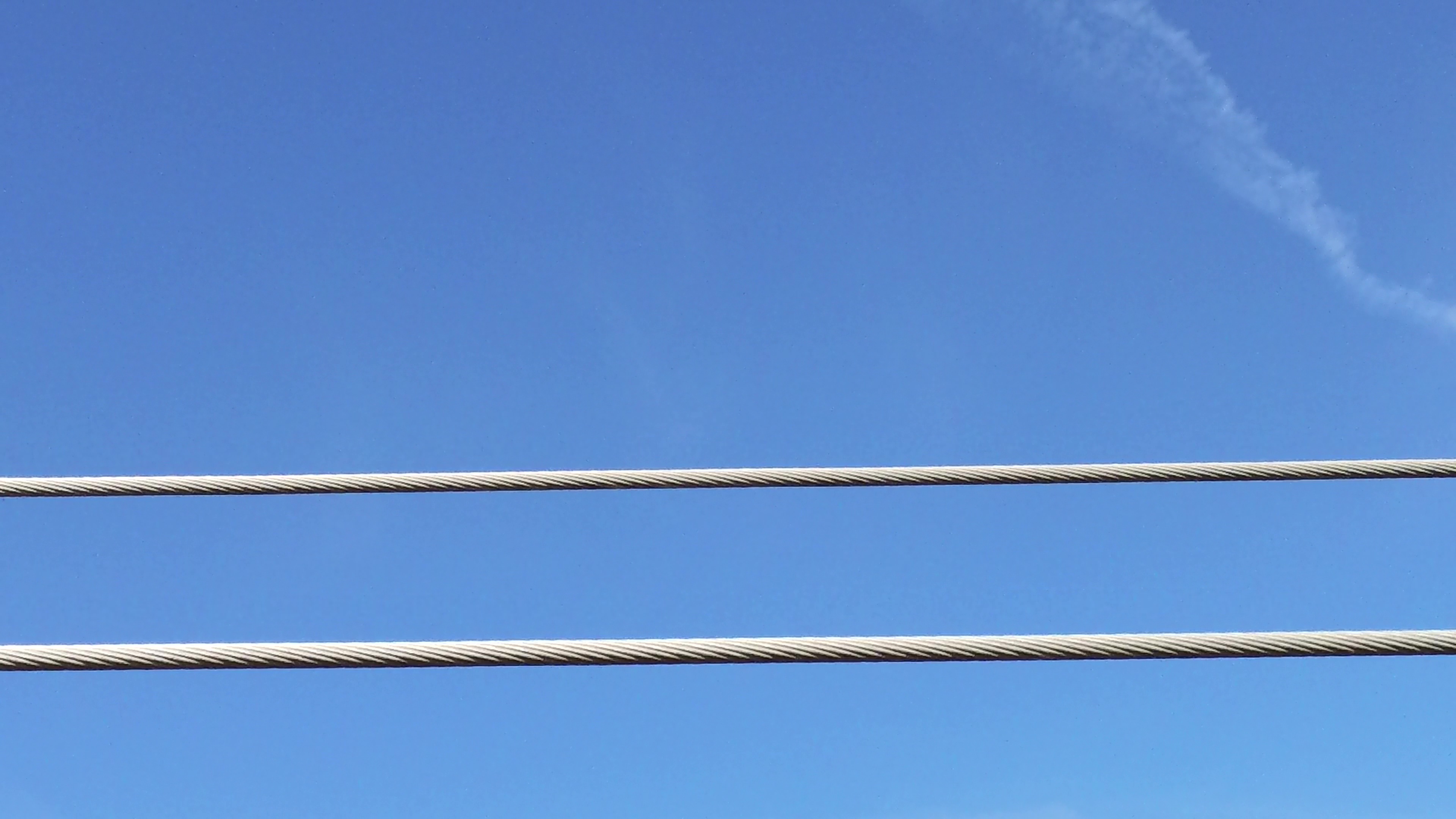}
		\vskip-2mm
	\caption{PLs in the TTPLA dataset.}
	\label{fig:powerline}
	\vskip-6mm
\end{figure}
\\ 	\\
\textbf{Segmentation Annotation.}
There are two types of segmentation: semantic segmentation and instance segmentation. Semantic segmentation assigns only one category label to all pixels that belong to the same class in a single image, while instance segmentation provides a mask at the pixel level to each individual instance in the image.  Because we must distinguish each individual PL and TT in UAV inspection, instance segmentation at the pixel level is desired for our dataset. To precisely label TTs and PLs, we use LabelME~\cite{russell2008labelme}. Each instance is surrounded carefully by a polygon.  Three expert annotators are recruited and in average, each person takes about 45 minutes to annotate one image. Each image is assigned to only one annotator to construct its full annotations. The annotation consistency between different annotators is actually not a serious issue in this work since 1) our images are mainly taken from a top view and therefore, we have very rare occlusions, 2) the instances in our datasets are well defined without much ambiguity based on our labeling policy, and 3) the three expert annotators label each assigned image with their highest possible scrutiny.  Samples of annotated images in TTPLA dataset are shown in Fig.~\ref{fig:mask}. 
\\ \\
\textbf{Labeling Instances.}
A new labeling policy is presented to categorize TTs based on lattice types and pole types (tubular steel, concrete, and wooden)~\cite{fang1999transmission}. 
\begin{itemize}
	\vskip-5mm
	\item 	Lattice TTs are composed of steel angle sections. TTPLA contains different shapes of lattice TTs ($T_{1}-T_{3}$) in Fig.~\ref{fig:towerShap} which are labeled by~``\textit{tower-lattice}''.
	\item Tubular steel, spun concrete, and steel/concrete hybrid poles belong to the same class.  These three types of poles have similar appearance. Our dataset contains three different shapes from this class ($T_{4}-T_{6}$) in Fig.~\ref{fig:towerShap}. To generate the label for this class, we take the first two letters from each type of poles and label such TTs as~``\textit{tower-tucohy}''.
	\item Wooden TTs have the poles made of wood. TTPLA considers this type of poles because wooden poles are distributed almost everywhere around residential places. So TTPLA contains a lot of different shapes of wooden poles such as $T_{7}$ and $T_{8}$ in Fig.~\ref{fig:towerShap}, which are labeled by~``\textit{tower-wooden}''.
	\vskip-5mm
\end{itemize}
The reason of labeling TTs in this way is to ensure that such a labeling policy is friendly to deep learning models. In general, each lattice tower can be divided into three parts: basic body (upper partition), body extension (middle partition), and leg extension (lower partition). Most lattice towers have similar shape in body extension and leg extension, which means that, if only body and/or leg extensions of two TTs appear in the image, it will be very hard for deep learning models to distinguish these two TTs.  This is also true for the TTs under ``\textit{tower-tucohy}''. Therefore, categorizing TTs based on their shapes (e.g., H-Frame, Lattice and monopole) may not be practical in UAV applications, since we cannot guarantee that UAVs always capture the basic body in the image.  To overcome this issue, TTs are categorized based on their structures and materials instead of their shapes in TTPLA dataset. Therefore, our labeling policy presents a good step toward the balance of the dataset.
Besides the labels related to TTs, two additional labels are presented:
\begin{itemize}
\vskip-9mm
	\item The label ``\textit{cable}'' is used for all PLs in TTPLA.
	\item The label ``\textit{void}'' is used for any instance (TT or PL) which is difficult to recognize into image. For example, a PL or even a TT may be labeled by ``\textit{void}'' if it is almost invisible in the image.  Any instances labeled by ``\textit{void}'' are ignored from evaluation~\cite{everingham2010pascal}. 
\end{itemize}
\vskip-9mm
\subsection{Dataset Statistics}
\label{s3:subsub4}
\begin{figure}[t]
	\centering
	\begin{tabular}{ c c c}
	\includegraphics[scale=0.45]{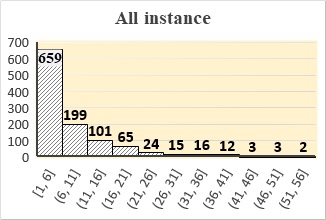} &
	\includegraphics[scale=0.45]{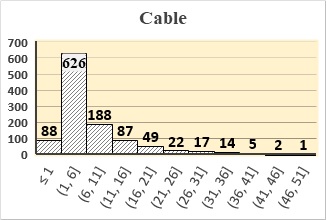} &
	\includegraphics[scale=0.45]{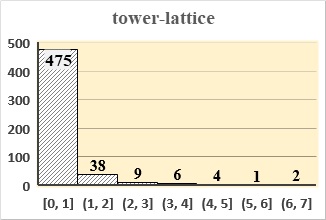} \\
	\includegraphics[scale=0.45]{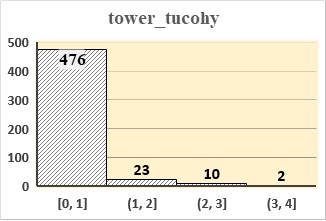}&
	\includegraphics[scale=0.45]{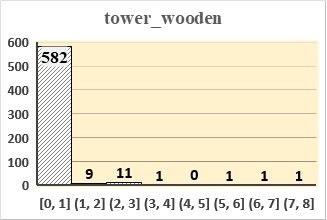} \\
	\end{tabular}
    \vskip-3mm
	\caption{Number of Instances per Image (x-axis) v.s. Number of Images (y-axis), (a) All instances, (b) \textit{cable}, (c)\textit{tower-lattice}, (d)\textit{tower-tucohy}, and (e)\textit{tower-wooden}.}
	\label{fig:statistic}
\vskip-3mm
\end{figure}

Fig.~\ref{fig:statistic} describes the relationship between the number of instances per image and the number of images. The left top corner figure demonstrates that there are 659 images that contain 1-6 instances per image and 241 images contains 11-56 instances per image. The others four figures describe the number of each specific object per image versus the number of images such as \textit{cable}, \textit{tower-lattice}, \textit{tower-tucohy}, and \textit{tower-wooden}. 
\begin{table}[b]%
	\vskip-5mm
	\caption {Dataset Statistics}
		\label{T:statis}\centering %
	\begin{tabular}{l | l | l | c | c |c }                           
		\toprule
		Category  &Classes &Labels  &Instances \#    &Instances/image   & Pixels\\
		\hline
		PLs         & Cable                  &\textit{cable} &8,083 &7.3   &154M \\
		\hline
		TTs   	    & Lattice                &\textit{tower-lattic} &330 &0.3 &164M    \\
		& Concrete/Steel/Hybrid              &\textit{tower-tucohy} &168 &0.15 &30M    \\
		& Wooden                             &\textit{tower-wooden} &283 &0.26  &61M  \\
		\hline
		       & Void                   &\textit{void} &173 &0.15 &0.8M \\
		\hline
	\end{tabular}
	\vskip-5mm
\end{table}

Statistics on the instances in TTPLA are reported in Table~\ref{T:statis}.  Notice that the number of instances in the cable class is much larger than those of the other classes on TTs. This is because a TT is always connected to at least 2 and up to 10 PLs.   Accordingly, TT classes have less training data. As a result, the numbers of instances on TTs and PLs will always be unbalanced in such a dataset focusing on individual instances. Although it is suggested in the Cityscapes dataset~\cite{cordts2016cityscapes} that the rare classes can be excluded from the evaluation, it should not be the case for our dataset since we are interested in a combination of both TTs and PLs.  In fact, to increase the number of TT instances in the dataset, we include images containing multiple TTs (see the figure at bottom-left in Fig.~\ref{fig:mask}), which are not often seen in other datasets.  An interesting observation is that the pixels that PLs and TTs occupied in the images are comparable, as reported in Table~\ref{T:statis}. It suggests that the dataset actually achieves a balance at the pixel level.  It would be interesting to investigate whether such a balance can benefit detection.
\section{Evaluation}\label{sec4}
This section presents metrics and loss functions that are used for training and evaluation. The baseline results are provided based on bounding boxes and instance masks.
	\vskip-9mm
\subsection{Metrics}
\label{E:first}
Instance segmentation on TTPLA is evaluated based on the standard metric of average precision (AP)~\cite{everingham2010pascal}. The intersection over union (IoU) measures the overlap between a pair of the matched prediction and the ground truth. Consequently, AP is accounted when the IoU is greater than 50\% \cite{hariharan2014simultaneous}. In the baseline, average precision is calculated for both bounding boxes, denoted by $AP_b$, and instance mask, denoted by $AP_m$~\cite{bolya2019yolact}.  Three precision scores are evaluated: $AP_b^{50\%}$, $AP_b^{75\%}$, $AP_b^{avg}$ for bounding box, and $AP_m^{50\%}$, $AP_m^{75\%}$, $AP_m^{avg}$ for masks~\cite{he2017mask,bolya2019yolact}, as listed in Table~\ref{results}.  $AP^{50\%}$ means the AP with the overlap value of 50\%, $AP^{75\%}$ means the AP with the overlap value of 75\%, and $AP^{avg}$ is the average AP value at different IoU thresholds ranging from 50\% to 95\% with step 5\% \cite{cordts2016cityscapes}. 
\subsection{Loss Function \label{E:third}}
Multi-loss functions are often used for multiple output networks~\cite{bolya2019yolact,he2017mask,liu2016ssd,girshick2015fast,redmon2016you} to measure the quality of prediction of each model's outputs by comparing them to the ground truth during the training process. In the baseline model, the multi-loss function $L_{loss}$ is a sum of localization loss $L_{loc}$, classification loss $L_{class}$, and mask loss $L_{mask}$, i.e., 
\vskip-3mm
\begin{equation} 
L_{loss}= \frac{\alpha}{N} L_{class} + \frac{\beta}{N} L_{loc} + \frac{\gamma}{A_{gb}} L_{mask}
\label{eq:loss_eqn}
\end{equation}
where $\alpha$, $\beta$, and $\gamma$ are the weights to balance the contribution of each loss function during the back-propagation. In the configurations, $\alpha$, $\beta$, and $\gamma$ are set to $1$, $1.5$, and $6.125$, respectively, similar to~\cite{bolya2019yolact}. In additions, $N$ is the number of boxes that matches the ground truth boxes. Moreover, the area of the ground truth bounding boxes $A_{gb}$ are used to normalize the mask loss.
\\\\
\textbf{Classification Loss Function.}  
With one label per bounding box, softmax loss function is used to estimate the confidence score $c_{i}^p$ of each proposed bounding box $i$ per category $p$, where $x_{ij}^p=\{0,1\}$ is the indicator of matching the $i$-th proposed box to the $j$-th ground truth box of category $p$, given $\hat c_{i}^p= \frac{\mathrm{exp}(c_{i}^p)}{\sum_{q}\mathrm{exp}(c_{i}^q)}$ 
\vskip-2mm
\begin{equation} 
\label{eq:class_eqn}
L_{class}(x,c)= - \sum_{i \in Pos}^{N} \sum_{j}  x_{ij}^p \mathrm{log} (\hat{c}_{i}^p) -\sum_{i \in Neg} \mathrm{log} (\hat{c}_{i}^p).\\
\end{equation}
\\
\textbf{Localization Loss Function.}  
Each bounding box has its center coordinate $(c_x,c_y)$, width~$w$, and height~$h$.  Smooth $L_1$ loss~\cite{girshick2015fast} is used to parameterize the bounding box offsets between the ground truth box $g$ and prediction box $l$.
\vskip-2mm
\begin{equation} 
L_{loc}(x,l,g)= - \sum_{i \in Pos}^{N}\sum_{j} \sum_{m \in c_x,c_y,w,h} x_{ij}^k \mathrm{smooth_{L1}} (l_{i}^{m}- \hat{g}_{j}^{m}).
\end{equation}
\\
\textbf{Mask Loss Function.}
Mask loss function is Binary Cross Entropy (BCE) loss between the predicted mask $M_{pr}$ and the ground truth mask $M_{gt}$ at the pixel level~\cite{bolya2019yolact}. Using BCE loss can maximize the accuracy of the estimated mask, where $L_{mask}= BCE(M_{gt},M_{pr})$.

\begin{table}[t]
\vskip-7mm
	\caption {Average Precision for Different Deep Learning Models on TTPLA.}
	\label{results}\centering %
	\begin{adjustbox}{width=0.9\textwidth}
	\begin{tabular}{c|l|cccccc}
		\toprule %
		Backbone  &Image size &$AP_b^{50\%}$  &$AP_m^{50\%}$ &$AP_b^{75\%}$  &$AP_m^{75\%}$  &$AP_b^{avg}$  &$AP_m^{avg}$\\
		\hline
		Resnet-50      & Yolact-640$\times$360 &46.72 &34.28 &4.99 &11.20 &16.50 &14.52  \\
		& Yolact-550$\times$550 &43.37 &28.36 &18.36 &12.22 &20.76 &14.70   \\
		& Yolact-700$\times$700 &42.62 &30.07 &20.36 &13.64 &21.90 &15.72   \\
		\hline
		Resnet-101     & Yolact-640$\times$360 &44.99 &32.58 &10.00 &10.06 &18.42 &14.05  \\
		& Yolact-550$\times$550 &45.30 &28.85  &19.80 &12.33 &22.61 &14.68    \\
		& Yolact-700$\times$700 &43.19 &28.18 &21.27 &13.46 &22.96 &14.88    \\
		\hline
	\end{tabular}
	\end{adjustbox}
	\vskip-6mm
\end{table}
\subsection{Baseline experiment results}
\label{E:third}

The images in TTPLA are split randomly into subsets of 70\%, 10\%, and 20\% images for training, validation, and testing, respectively. Yolact with different backbones are evaluated based on the proposed dataset. Yolact produces bounding box, confidence score for a true object, and mask for each predicted object instance.  Yolact is trained based on our dataset using two GeFoce GTX-1070 GPU with 8G memory/each. We train the model using different image sizes $640\times360$ (preserve aspect ratio), $550\times550$ and $700\times700$. In addition, different backbones are used in our training such as resnet-101 and resnet-50.  All the results on average precision are reported in Table~\ref{results}. The best average scores for bounding box and mask are $22.96\%$ and $15.72\%$, respectively, as listed in Table \ref{results}. Overall, the average precision for instance mask level is less than that of bounding box. 

\begin{wrapfigure}{r}{0.38\textwidth}
        \vskip-8mm
		\includegraphics[scale=0.33]{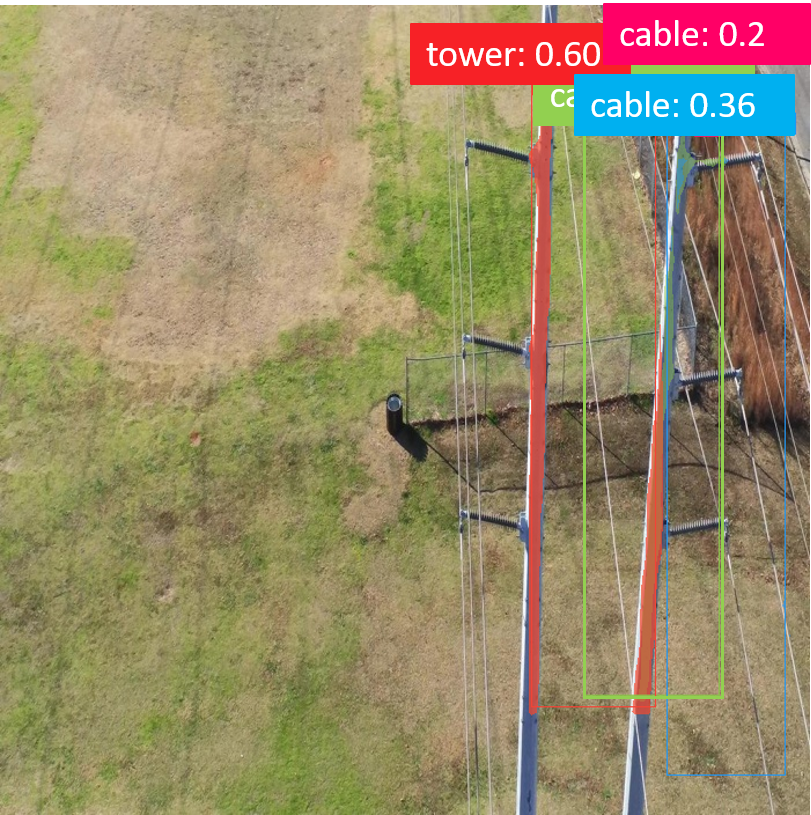}
		\captionof{figure}{Classification falseness.}
		\label{fig:classification}
		\vskip-8mm
		\end{wrapfigure}
A brief case study is presented as follows based on the TTPLA dataset. Average precision is evaluated based on true positives and false positives. False positive is considered for any object with IoU less than 50\%. In addition, there is an inversely proportional relation between average precision and the number of false positives.  Therefore, false positive increases as a result of three types of falseness on classification, detection and segmentation.

Firstly,  classification falseness appears as a result of confusion on class labels. Although there is no shape similarity between the classes of PLs and TTs, there is still a small proportion of this type of~falseness which is up to 1.3\% from the test-set images. Further examination of the results show that the classifier may not be able to distinguish one type of TTs (\textit{tower-tucohy}) and PLs as shown in Fig.~\ref{fig:classification}. One possibility of this confusion is that the color and shape of small-size \textit{tower-tucohy} have high similarity as those of PLs. This type of falseness is considered as a challenge and leaves much scope for improvement.

\begin{wrapfigure}{r}{0.69\textwidth}
		\vskip-9mm
		\includegraphics[scale=0.27]{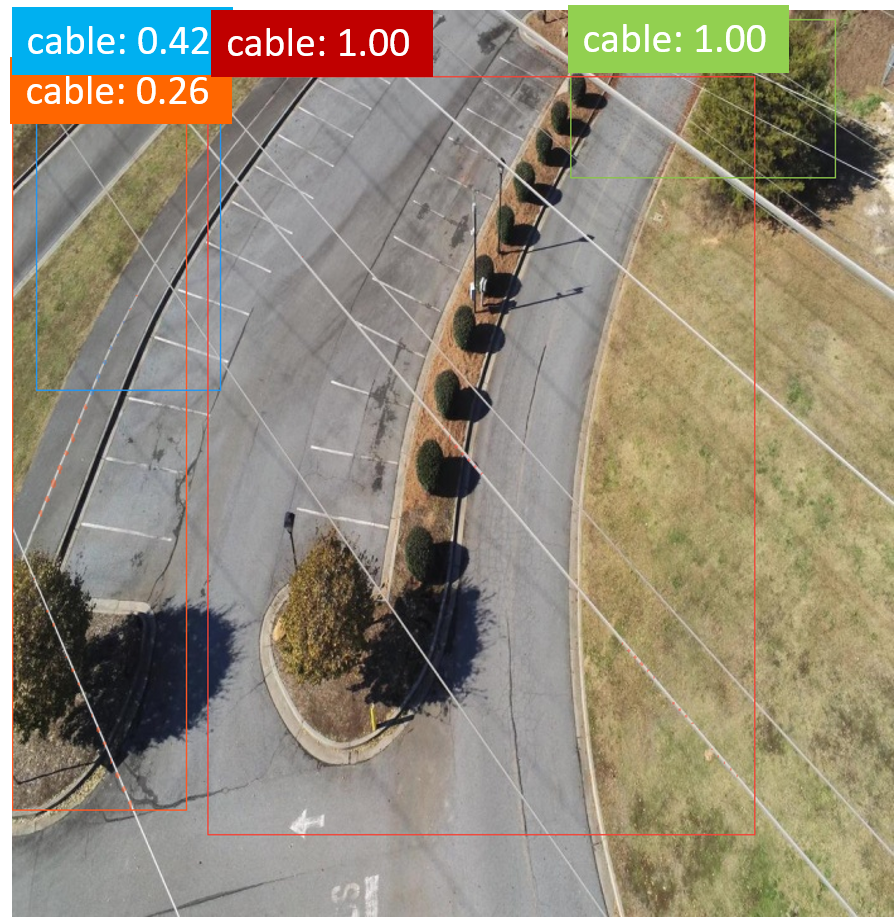}
 		\includegraphics[scale=0.27]{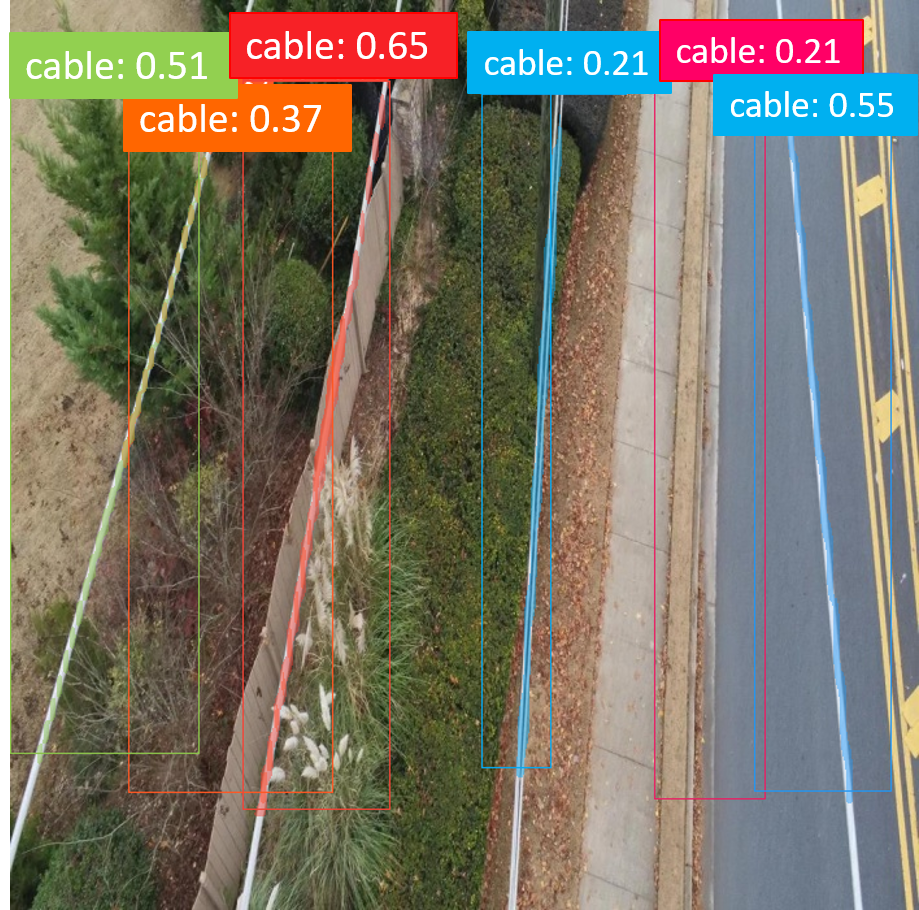}
				\vskip-2mm
		\captionof{figure}{Detection falseness.}
		\label{fig:detect}
		\vskip-8mm
\end{wrapfigure}
Secondly, detection falseness is produced due to one of the following two reasons. On one hand, the object is not detected. On the other hand, an object may be detected in a region where there is actually no object. As shown in Fig.~\ref{fig:detect}, there is a wrong detection of PL in regions of lane line and sidewalk, respectively. Based on what is mentioned in \cite{everingham2010pascal}, the probability of detection falseness is high similar because PLs do not have predictable visual properties and all PLs have the same features without much distinction. On the other hand, significant shape variation of TTs affects directly the precision of detection. To reflect this point in our dataset, as mentioned in subsection~\ref{s3:sub2}, we collect images for TTs from different views.  

\begin{wrapfigure}{r}{0.4\textwidth}
        \vskip-8mm
		\includegraphics[scale=0.3]{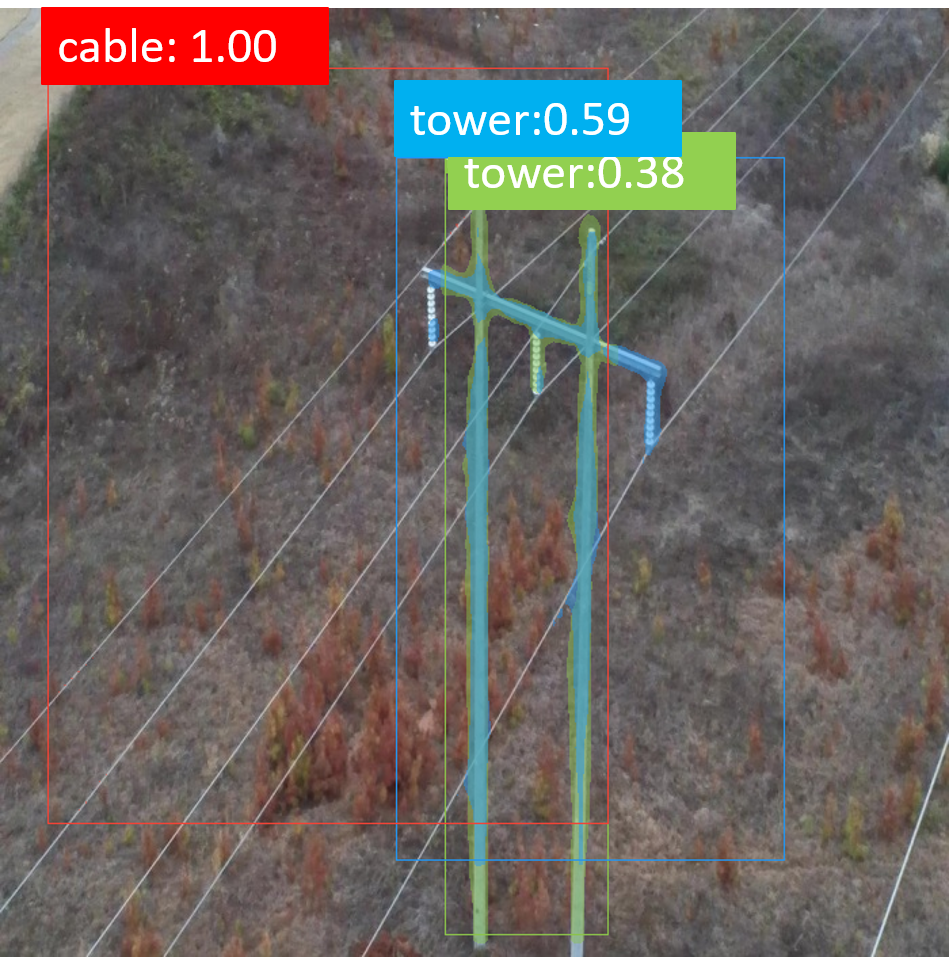}
				\vskip-2mm
		\captionof{figure}{Segmentation falseness.}
		\label{fig:segmentation}
		\vskip-7mm
		\end{wrapfigure}  
Thirdly, segmentation falseness appears when the segmentation mask is not covering the whole object. As mentioned in \cite{everingham2010pascal}, there is a strong relationship between precision and the object size.  In other words, precision can be improved when the number of object pixels increases. This is due to the difficulty of extracting the feature of small objects specially with noisy background. This problem often appears in detecting PLs, because of their long-thin shape and simple appearance. In TTPLA, most PLs have very small width between 1 to 3 pixels. In addition, PLs are so long as compared to the size of images, when compared to the instance objects included in COCO and PASCAL VOC. Consequently, according to Yolact, PLs are detected by only one bounding box and only one mask which in most cases is not covering the whole  PL and leads to reduced mask average precision for PLs. Moreover, in most cases the single power line is split up to multiple detecting instances which also increases the falseness for segmentation \cite{de2017semantic}. This falseness also appears with TTs detection as shown in Fig. \ref{fig:segmentation}.

Fourthly, NMS is exploited by instance segmentation detectors ~\cite{bolya2019yolact,he2017mask}, that produce large numbers of false positives near the ground truth, to suppress the overlapped bounding boxes based on lower confidence score and the overlap threshold \cite{liu2019adaptive}.  In the crowded scenario, the objects are quite close, overlapped and their predicted bounding boxes can overlap with each other. Therefore, some bounding boxes are suppressed based on overlap threshold of NMS although its nearby bounding boxes are actually for different objects, which reduces the average precision. Changing the overlap threshold may be one solution, however in the crowded scenario it is not a perfect solution since higher NMS threshold leads to increased false positives while lower NMS threshold may increase the miss-rate and remove more true positives \cite{liu2019adaptive}. The number of the overlapped bounding boxes per object is reported in Table \ref{T:statis1}. The overlap is calculated based on threshold 30\%, 50\%, 75\% and 95\%, respectively. For example, we have 4,251 overlapped bounding boxes of PLs with threshold 30\%. As reported in Table \ref{T:statis1}, in TTPLA dataset, the total percentage of the overlap between the bounding boxes of different instances is up to 48.9\%, 36.8\%, 17.9\%, 2.5\% for threshold 0.3, 0.5, 0.75, 0.95, respectively. 

Finally, the analysis results highlight the difficulties to process these real-time images collected by autonomous~UAV and reflect the challenges included in our dataset which pose opportunities for further enhancements.
\begin{table}[t]
		\vskip-6mm
		\caption{Total Percentage of Overlap on TTPLA.}
		\label{T:statis1}
		\centering
		\begin{tabular}{lcccc}
			\toprule
			Category  & Overlap (30\%) & Overlap (50\%) &Overlap (75\%)   &Overlap (95\%)\\
					\hline
		\textit{cable} &4,251 &3,224 &1,570 &224\\
		\textit{tower-lattice}     & 15   &3 &0 &0    \\
		\textit{tower-tucohy}      & 20    &4 &2 &0 \\
		\textit{tower-wooden}      & 22     &10 &2 &0 \\
		Total(\%)                  &48.9 &36.8 &17.9 &2.5 \\
			\bottomrule
		\end{tabular}
\vskip-5mm
\end{table}
\section{Conclusion\label{sec5}}
TTPLA is the first public image dataset with a focus on combined TTs and PLs instance segmentation. TTPLA dataset consists of 1,100 aerial images with resolution of 3,840$\times$2,160 and contains up to 8,987 instances. Data collection and labeling for TTPLA dataset are highly challenging to ensure the variety in terms of view angles, scales, backgrounds, lighting conditions and zooming levels. Therefore, novel policies are proposed for collecting, annotating, and labeling the aerial images.  TTPLA dataset is annotated accurately at the pixel-wise level to be employed by the instance segmentation using deep learning models. Based on TTPLA, a baseline is created using the state-of-the-art learning model, different backbones, and various images sizes. Finally, TTPLA dataset can provide a new challenge to computer vision community and lead to new advancement in detection, classification and instance segmentation.
\\
\\
\textbf{Acknowledgments.}
The authors gratefully acknowledge the partial financial support of the National Science Foundation (1830512).

\bibliographystyle{splncs}
\bibliography{egbib}

\end{document}